%% file: paper_template.tex
\documentclass[conference]{IEEEtran}
\usepackage{times}

\usepackage[numbers]{natbib}
\usepackage{multicol}
\usepackage[bookmarks=true]{hyperref}
\usepackage{amsmath}

\usepackage{pifont}
\usepackage{xcolor}
\usepackage{amsmath} 
\usepackage{amssymb}  
\usepackage{amsthm}
\usepackage{algpseudocode}
\usepackage{xcolor}
\usepackage{lipsum}
\usepackage{booktabs}
\usepackage{adjustbox}
\usepackage{mathtools}
\usepackage{float}
\usepackage{nicefrac, xfrac}
\usepackage{arydshln}
\usepackage{mathtools}
\usepackage{yhmath}
\usepackage{soul}

\newcommand{\algname}{JIGGLE}
\newcommand{\thinlayer}{{\mathcal{T}}}

\newcommand{\bel}{{\mathbf{b}}} 
\newcommand{\action}{{u}} 
\newcommand{\pbd}{{f}} 
\newcommand{\mub}{{\widehat{\mathbf{b}}}} 
\newcommand{\Sigmab}{{\boldsymbol\Sigma}} 
\newcommand{\statesim}{{\mathbf{x}}} 
\newcommand{\stateref}{{\mathbf{x}^{\text{ref}}_t}}

\newcommand{\staterefp}{{\mathbf{x}^{\text{ref}}_{t+1}}}
\newcommand{\motionmodel}{{m}} 
\newcommand{\motionnoise}{{w}} 
\newcommand{\motioncov}{{W}} 
\newcommand{\obsmodel}{{h}} 
\newcommand{\beljacob}{{J}} 
\newcommand{\obsnoise}{{v}} 
\newcommand{\noisejacob}{{R}} 
\newcommand{\residualcovariance}{{S}} 
\newcommand{\obscov}{{V}} 
\newcommand{\kalmangain}{{K}} 

\newcommand{\pcl}{{Q}} 
\newcommand{\upgreenarrow}{{\rotatebox{90}{\color{green!80!black}\ding{225}}}}

\newcommand{\entropy}{{\mathcal{H}}} 
\newcommand{\energy}{\mathcal{U}} 
\newcommand{\boundaryenergy}{\energy_{b}} 
\newcommand{\inworkspace}{{\Omega}} 
\newcommand{\weighteddeform}{{\mathcal{D}}} 

\newcommand{\boundarypt}{{\mathbf{x}^{\mathsf{B}}_{t+1}}}
\newcommand{\otherpt}{{\mathbf{x}^{\mathsf{U}}_{t+1}}}
\newcommand{\boundaryset}{{\mathbf{x}^{\mathsf{B}}_{t+1}}}
\newcommand{\otherset}{{\mathbf{x}^{\mathsf{U}}_{t+1}}}
\newcommand{\singularval}{{\lambda_{i}}}

\definecolor{rebuttal}{rgb}{0, 0, 0 }
\definecolor{notecolor}{rgb}{0.188, 0.624, 0.478}
\definecolor{feicolor}{rgb}{0.188, 0.478, 0.624}
\definecolor{feicolorhighlight}{rgb}{0.816, 0.282, 0.282}

\makeatletter
\newcommand*\MY@leftharpoonupfill@{\arrowfill@\leftharpoonup\relbar\relbar}
\newcommand*\MY@rightharpoonupfill@{\arrowfill@\relbar\relbar\rightharpoonup}
\newcommand*\overleftharpoon{\mathpalette{\overarrow@\MY@leftharpoonupfill@}}
\newcommand*\overrightharpoon{\mathpalette{\overarrow@\MY@rightharpoonupfill@}}
\makeatother

\newcommand{\norm}[1]{\left\lVert#1\right\rVert}

\newtheorem{theorem}{Proposition}

\newenvironment{proposition}{\begin{theorem}%
  \pushQED{\qed}}%
  {\popQED\end{theorem}}

\newtheorem{theorem2}{Lemma}
\newenvironment{lemma}{\begin{theorem2}%
  \pushQED{\qed}}%
  {\popQED\end{theorem2}}

\usepackage{etoolbox} 
\AtBeginEnvironment{pmatrix}{\everymath{\displaystyle}}
\AtBeginEnvironment{bmatrix}{\everymath{\displaystyle}}

\let\labelindent\relax
\usepackage{enumitem}

\setlist[itemize]{leftmargin=10pt}
\definecolor{ForestGreen}{rgb}{0.133, 0.545, 0.133}
\usepackage[ruled,vlined,linesnumbered]{algorithm2e}

\SetCommentSty{mycommfont}
\begin{document}


\title{
\textbf{JIGGLE}: An Active Sensing Framework for Boundary Parameters Estimation in Deformable Surgical Environments}
\author{Nikhil Uday Shinde$^*$, Xiao Liang$^*$, Fei Liu\IEEEmembership{Member, IEEE}, Yutong Zhang, Florian Richter, Sylvia Herbert and Michael C. Yip\IEEEmembership{Senior Member, IEEE}
\\
$^*$ \textit{These authors contributed equally to this work}
\\
University of California San Diego
\\
Email: \{nshinde, x5liang, f4liu, yuz049, frichter, sherbert, yip\}@ucsd.edu
}



%

\maketitle

\begin{abstract}
Surgical automation can improve the accessibility and consistency of life-saving procedures. 
Most surgeries require separating layers of tissue to access the surgical site, and suturing to re-attach incisions. 
These tasks involve deformable manipulation to safely identify and alter tissue attachment (boundary) topology. 
Due to poor visual acuity and frequent occlusions, surgeons tend to carefully manipulate the tissue in ways that enable inference of the tissue’s  attachment points without causing unsafe tearing. 
In a similar fashion, we propose JIGGLE, a framework for estimation and interactive sensing of unknown boundary parameters in deformable surgical environments. 
This framework has two key components: 
(1) a probabilistic estimation to identify the current attachment points, achieved by integrating a differentiable soft-body simulator with an extended Kalman filter (EKF), and
(2) an optimization-based active control pipeline that generates actions to maximize information gain of the tissue attachments,  
while simultaneously minimizing safety costs.
The robustness of our estimation approach is demonstrated through experiments with real animal tissue, where we infer sutured attachment points using stereo endoscope observations. 
We also demonstrate the capabilities of our method in handling complex topological changes such as cutting and suturing.

\end{abstract}

\IEEEpeerreviewmaketitle

\section{Introduction}\label{sec:introduction}
\input{main_tex_files/introduction}

\section{Related Works}\label{sec:related_works}
\input{main_tex_files/related_works}

\section{Problem Statement}\label{sec:problem_statement}
\input{main_tex_files/problem_statement}

\section{Methods}\label{sec:methods}

\input{main_tex_files/methods}

\section{Evaluations}\label{sec:evaluations}
\input{main_tex_files/evaluations}

\section{Conclusion}\label{sec:conclusion}
\input{main_tex_files/discussion_conclusion}
\bibliographystyle{plainnat}
\bibliography{references}

\clearpage
\input{appendix}

\end{document}

%% file: main_tex_files/introduction.tex

\begin{figure}[t]
    \centering
    \includegraphics[width=0.98\linewidth]{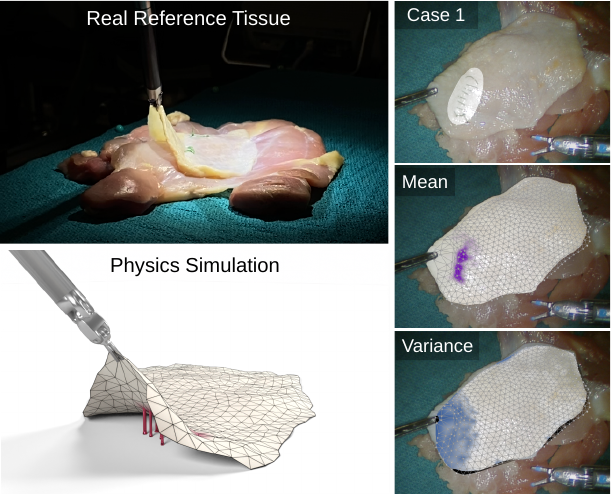} 
    \caption{
    JIGGLE conducts probabilistic estimation of soft tissue attachment points from image data and manipulation of the tissue.
    In our real-world ex-vivo experiments, a surgical tool manipulates chicken skin that is attached to its environment with sutures, and we use a stereo endoscope as the observation to estimate the boundary parameters in a probabilistic fashion.
    The estimated boundary, i.e. the suture locations, is shown in purple.
    Since the boundary parameters are estimated probabilistically, a corresponding confidence metric can be found and is shown in blue.
    Note that the confidence is only high near the grasping location because that is where the deformation is applied.
    }
    \label{fig:cover_figure}\vspace{-1.5em}
\end{figure}

Surgical automation has the potential to improve the accessibility of life-saving procedures in under-served communities. 
Automation can help increase consistency between patient outcomes and alleviate the load of routine procedures from a burdened healthcare system. 
In recent years, the robotics community has made large strides in surgical automation through the development of dVRK \cite{6907809}, works like STAR \cite{STAR}, and improvements in the automation of tasks such as \textcolor{rebuttal}{deformable objects manipulation \cite{hu20193, murphy2023surgical, thach2022learning}}, suturing \cite{6dofsuture,autosuture}, debridement \cite{debridement}, blood suction \cite{suction}, cutting \cite{cutting} and \textcolor{rebuttal}{disection \cite{10354422}}.

Deformable manipulation is a significant component of all these surgical tasks. In fact, large portions of a surgical procedures (and occasionally, its majority) involve safely cutting tissue in order to detach regions and access the site of the surgery, as well as suturing tissue back together. 
These actions actively deform and change the structure of the surgical scene. 
A key step towards realizing surgical autonomy is enabling robots to understand and track these changing structures. 
Many previous works have aimed to take this step solely through 3D scene reconstruction \cite{SUPER, lin2023semanticsuper, edsrr, wang2022neural}. However, these works are insufficient to explore safe interaction with the environment as they do not provide an estimate of how underlying tissue structures are being deformed or provide an understanding of how aggressively the tissue is being manipulated. 


In order to move towards safer surgical autonomy, an interactive approach to track the tissue and its structure is desirable because it can lead to more intelligent tissue control. 
%
Previous approaches like \cite{boonvisut2014identification} that do consider the joint problem of estimation and active sensing are limited in the resolution of attachment regions they can detect. Safety is not considered, and behaviors are limited to hand-tuned motion primitives.
Additionally, previous works do not show their results on real tissue data and fail to consider topological changes like cutting and suturing. 

In this work, our goal is to realize an interactive approach to estimating, manipulation, and tracking of a deformable \textcolor{rebuttal}{\st{3D} thin-shell} tissue in such a way that we can reason about its structure and apply more intelligent, safer tissue control. Specifically, we propose the Joint Interactive Guided Gaussian Likelihood Estimation (\algname) method: a novel active sensing framework for estimating boundary attachment points in deformable surgical environments. 
We leverage a differentiable physics simulator to develop an efficient probabilistic estimation framework for deformable environments with a high degree of freedom (DOF).
Our estimation framework relies solely on stereo camera observations which is typically the only available feedback modality for interactions with the environment.
We demonstrate the robustness of this estimation in real tissue manipulation experiments, as shown in Fig.~\ref{fig:cover_figure}. 
We take advantage of the probabilistic nature of the estimation framework to build an active sensing pipeline that selects actions to maximize information gain while minimizing a safety cost to avoid tissue tearing. 
We showcase the complete \algname~framework in simulations involving topological changes like cutting and suturing that actively change the underlying attachment points. 
We summarize our contributions as the following: 
\begin{itemize}
    \item An online probabilistic estimation framework to track attachment points in high DOF, deformable environments
    \item An active sensing algorithm with fine control to maximize information gain while respecting safety constraints
    \item Demonstrating our framework with topological changes like suturing and cutting. 
    \item Real-world experiments on animal tissue showcasing our robust estimation framework and simulation experiments demonstrating and analyzing the performance of our estimation and active sensing methods. 
\end{itemize}

%% file: main_tex_files/related_works.tex
    \textbf{Active Sensing.} 
    In robotics, active sensing involves selecting appropriate actions for a robot to maximize information gain while minimizing costs \cite{mihaylova2003comparison}.
    An active sensing framework contains an estimator (e.g. a Bayesian filter) that can estimate the system's state under uncertainty.
    Using this state estimation, an objective relating to information gain is maximized. 
    Information theoretic objectives can be specified in terms of entropy \cite{greigarn2019task}, conditional entropy \cite{RYAN2010574}, and mutual information \cite{jadidi2015mutual, asgharivaskasi2022active}. 
    To optimize information gain, frameworks such as trajectory optimization \cite{6760585} or partially observable Markov decision processes (POMDPs) \cite{sunberg2018online} can be applied to determine the next robot actions. 
    Previously, these techniques have been applied extensively to simultaneous localization and mapping (SLAM) \cite{6580252, asgharivaskasi2022active}, multi-agent tasks \cite{8260881}, and robot manipulation \cite{schneider2022active}.
    Another application of active sensing is tissue stiffness discovery for tumor localization using palpation.
    Previous work on this application focused on guiding the sampling process of Gaussian processes (GPs) to represent variable tissue stiffness \cite{9086901, garg2016tumor, 8460936}.
    
    \textbf{Differentiable Physics Simulator.} 
    Differentiable simulators enable easy calculation of multiple orders of derivatives of outputs.
    This characteristic is especially useful in robotic applications with deformable objects as it enables efficient optimization for manipulation and control. 
    The authors of \cite{Du_2021_diffpd, diffcloth} built a differentiable simulation based on projective dynamics (PD) and showed successes in parameter identification of soft bodies and optimal control tasks involving cloth-like objects.
    Position-based dynamics (PBD) \cite{macklin2016xpbd, muller2007position} is another favored technique, given its speed and flexibility. 
    It has been used for scenarios such as the manipulation of deformable linear objects (DLOs) \cite{Fei_2023_DLOs_RAL}, safe manipulation of cloth \cite{zhang2023achieving}, and material parameter optimization \cite{10.1145/3606923, chen2022virtual}.
     Other popular simulation frameworks, such as the finite element method (FEM) and the material point method (MPM), have also been applied widely in robotic tasks \cite{heiden2021disect, 9720966, hu2019difftaichi}.
     The surgical robotics community has recently begun researching the use of differentiable simulators in complex surgical scenarios. 
     Specifically, \cite{9561177, liang2023realtosim} identifies and reduces the gap between real and simulated tissue via online parameter estimation. 
     
     In this work, we leverage a differentiable extended PBD (XPBD) simulator. 
     We formulate a new type of geometric constraint that represents the attachment of a tissue to its environment. 
     We rely on the simulator's differentiability to perform gradient-based updates on the constraints' parameters.

    \textbf{Boundary Condition Identification.}
    Besides stiffness estimation, understanding attachment points and boundary parameters has also been explored. 
    In \cite{peterlik2014model}, the authors use deformable registration of a simulation to CT images to determine boundary parameters from the forces calculated on the simulated mesh. 
    The authors in \cite{xu2022identification} formulates boundary conditions as virtual springs and estimate them using an interior point solver, but do \textcolor{rebuttal}{not} consider active sensing.
    Other approaches take a fully perception-based approach to understanding boundaries in surgical environments.
    Semantic segmentation based approaches like \cite{lin2023semanticsuper} estimate visible boundaries directly from endoscopic data. 
    For attachments that are not directly visible, \cite{9359348, tagliabue2021intra}, developed a deep learning based method to directly predict attachment points of soft tissues. 
    However, they fail to consider active sensing; instead, they take a deterministic single-shot approach that fails to consider probabilistic estimation. 
    This confidence information is crucial for tracking boundary estimates and understanding information gain, which is vital to active sensing. 
     \cite{boonvisut2014identification} takes initial steps in solving the joint problem of estimation and active sensing. 
    Their estimation approach relies on a coarsely discretized set of possible attachment points with binary connections to the environment. 
    This assumption inhibits their method from scaling to higher DOF models and capturing complex attachment regions.
    Their active sensing approach relies on an exhaustive evaluation of pre-defined motion primitives. 
    Hand-designing motion primitives is not generalizable and incapable of generating new complex deformations required for identification. 
    Additionally, they do not account for safety in their approach. 
    We compare against their active sensing approach in our evaluations. 
    Finally, \textcolor{rebuttal}{ that} work fails to consider topological changes like cutting and suturing.

%% file: main_tex_files/problem_statement.tex
In this paper, we consider a deformable thin shell layer $\thinlayer$ in an environment in $\mathbb{R}^{3}$. 
$\thinlayer$ represents a tissue in the context of surgery \textcolor{rebuttal}{such as skins and membranes} or may be used to represent deformable objects like cloth for household tasks. 
In this paper, we will refer to the thin shell as a tissue. 
We assume the tissue $\thinlayer$ has non-overlapping subsets $\thinlayer_{f}$ and $\thinlayer_{b}$ such that $\thinlayer = \thinlayer_{f} \cup \thinlayer_{b}$. 
$\thinlayer_{f}$ represents the regions of the tissue that are free to move. 
$\thinlayer_{b}$ represents regions whose movement is constrained by the environment (i.e. boundary constraints). 
In a surgical environment, these constraints may be the result of natural tissue connections, such as fascia or wound closures. 
We refer to these boundary constraints on the tissue as \textit{attachment points}, and we assume their strength can vary spatially across the tissue. 
In addition, the attachment points can change over time due to actions such as cutting and suturing, that actively alter the tissue's attachments to the environment, during surgery. 
We assume that the tissue is manipulated using a position-controlled robot end-effector.
The robot grasps and directly moves a sub-region of the free tissue surface $\thinlayer_{g} \subset \thinlayer_{f}$.

Without loss of generality, we use a thin shell mesh to discretely represent the tissue, $\thinlayer$. 
This mesh is composed of particles that are locally connected by edges. 
The state of the tissue at time $t$ is specified by its $n$ particle positions $\statesim_{t} = [\statesim^{1}_{t}, \dots, \statesim^{n}_{t}]^\top,$ where $\statesim^{i}_{t} \in \mathbb{R}^{3}, \forall i \in [1, \dots, n]$ specifies the position of particle $i$. 
We denote the position-based grasp control of $\thinlayer$ as $\action_{t} \in \mathbb{R}^{3}$. 
$\bel_{t} = [b^{1}_{t}, b^{2}_{t} \dots, b^{n}_{t}]^\top \in \mathbb{R}^{n \times 1}$ is a parametric representation of the boundary constraints on tissue $\thinlayer$.
In our work, we select each $b^{i}$ to signify the strength of the environmental attachment constraint for particle $i$.
The exact form of these attachment constraints can be tailored to the method chosen to solve the problem statement and hence is discussed later in section \ref{sec:PBD_simulator}.
By parameterizing the attachment constraints for the entire tissue, $\bel_{t}$ fully specifies the topology and structure of $\thinlayer_{f}$ and $\thinlayer_{b}$. 
We will refer to this parameter $\bel$ as the boundary parameter. 

In this paper, we consider the following estimation task. 
We assume there exists a real tissue $\thinlayer^{\text{ref}}$ with its attachment constraints specified by true parameters $\bel^{*}_{t}$. 
We have access to observations on the state of the reference tissue: $\stateref$.
Our goal is to estimate the true parameters, $\bel^{*}_{t}$, using the observed state of the tissue, $\stateref$, and $\bel_{t}$ denotes our estimated belief.

In the active sensing problem, we aim to find the best actions 
\textcolor{rebuttal}{to} estimate $\bel_{t}$. 
The robot must first solve the estimation problem using observations from its current action to update its estimate $\bel_{t}$. 
Then, using this estimate, it must decide on the next action, $\action_{t+1}$, to maximize future information gain on parameter $\bel_{t}$. 
To formalize this, we consider our estimate $\bel_{t}$ of the true parameters $\bel^{*}_{t}$ as a random variable and seek a controller that minimizes the entropy of our predicted belief, $\entropy(\bel_{t+1})$.

\begin{algorithm}[!t]
\small
    \caption{JIGGLE: online active sensing framework for estimating boundary constraints. }
    \label{alg:full_algorithm}
    \SetKwInOut{Input}{Input}
    \SetKwInOut{Output}{Output}

    \Input{\text{Initial particles} $\mathbf{x}_0$, \text{Initial boundary estimation} $\mathbf{b}_0$, initial covariance $\Sigmab_0$, \text{real-world tissue }$\mathcal{T}^\text{ref}$ or ground truth parameters $\bel^\text{ref}$, initial control point $u_0$, \text{motion noise covariance} $W$, horizon $T$}

        $\mub_0, \Sigmab_0 \gets \bel_0, \Sigmab_0$

        $t_c \gets 0$
        
        \For{t in $[0,...,T]$}{

            \tcp{Incorporate topological changes}
            
            \If {topological change} {
                $t_c = t$

                \If {in simulation} {
                                
                $\mathbf{b}^\text{ref}, \delta \bel_t, W_t \gets \text{PerformCutOrSuture}(\mathbf{b}^\text{ref})$
                } \Else {
                
                $\mathcal{T}^\text{ref} , \delta \bel_t, W_t \gets \text{PerformCutOrSuture}(\mathcal{T}^\text{ref})$
                }
            } \Else {
                $\delta \bel_t, W_t \gets \mathbf{0}, W$
            }

            $\displaystyle\mub_{t+1|t}, \Sigmab_{t+1|t} \gets \text{EKFPredict}(\mub_{t|t}, \Sigmab_{t|t}, \delta \bel_t, W_t)$

            \tcp{Estimate parameters}
            
            $M \gets \text{SampleTimeStep}(t_c, t)$

            \If {in simulation}{
                 $\displaystyle \stateref \gets f(\stateref, u_t, \bel^\text{ref})$
                 
                $\displaystyle \mub_{t+1}, \Sigmab_{t+1} \gets \text{EKFUpdate}(\mub_{t+1|t}, \Sigmab_{t+1|t}, \{\mathbf{x}^\text{ref}_{t'}\}_{t'=t_c}^{t}, M)$
            } \Else {
                $\mathcal{T}^\text{ref}\gets \text{ApplyControl}(\mathcal{T}^\text{ref}, u_t)$
            
                $Q_t \gets \text{Perception}(\mathcal{T}^\text{ref})$
                
                $\displaystyle \mub_{t+1}, \Sigmab_{t+1}, \stateref \gets \text{JointEst}(\mub_{t+1|t}, \Sigmab_{t+1|t}, \{\mathbf{x}^\text{ref}_{t'}\}_{t'=t_c}^{t}, \{Q_{t'}\}_{t'=t_c}^{t}, M)$
            }

            \tcp{Decide the next action}
            \If {active sensing} {
                $\delta u \gets \text{ActiveSensing}(\mathbf{x}^{\text{ref}}_t, u_t, \mub_{t+1}, \Sigmab_{t+1})$

            } \Else {
                $\delta u \gets \text{UserDefine()}$
                
            }

            $u_{t+1} =  u_t + \delta u$
        }
        \vspace{0.5em}
\end{algorithm}

%% file: main_tex_files/methods.tex
This section covers our approach to \textcolor{rebuttal}{1) estimation, and 2) active sensing} of boundary parameters. 
We first describe our simulation environment in \ref{sec:PBD_simulator}, and how it is used for estimation in \ref{sec:estimation_framework}. 
We discuss extensions of this framework, necessary for use with real world data in \ref{sec:joint_estimation}. 
\ref{sec:active_sensing} describes the objectives and constraints used for active sensing, followed by the algorithms used to solve for control in \ref{sec:active_exploration_experiment}.

\subsection{Deformable Softbody Simulator} \label{sec:PBD_simulator}

\textbf{Extended Position-based Dynamics (XPBD). } 
In this work, we model the problem using a \textit{quasi-static} extended position-based dynamics (XPBD) simulator. 
XPBD was chosen for its computational efficiency and ease of specifying new geometric constraints. The following paragraph is a brief summary of the XPBD method and update rule~\cite{macklin2016xpbd}.

At time $t$, the robot end effector chooses its next position as its action. 
This is denoted as $\action_{t}\in\mathbb{R}^3$. 
Let $\statesim_{t}$ be the tissue mesh particle states at time $t$.
XPBD solves for the next state, $\statesim_{t+1}$, by minimizing the total potential energy of the system: 
\begin{equation} \label{eqn:energy_minimization_xpbd}
    \statesim_{t+1} = \arg\min_{\statesim_{t+1}}\energy(\statesim_{t+1}, \action_{t})
\end{equation}
The potential energy is defined as: 
\begin{equation}\label{eqn:elastic_potential}
    \begin{split}
        \energy(\statesim_{t+1}, \action_{t}) = \large \sfrac{1}{2}\ \mathbf{C}(\statesim_{t+1}, \action_{t})^{\top}\mathtt{diag}(\mathbf{k})\mathbf{C}(\statesim_{t+1}, \action_{t})
    \end{split}
\end{equation}
Here, $\mathbf{C}(\statesim_{t+1}, \action_{t})$ is the union ($\cup$) of all the position-based constraints that define the system. 
The simulator aims to satisfy these constraints through energy minimization. 
$\mathbf{k}$ is a vector that defines the relative weighting on the constraints in $\mathbf{C}(\statesim_{t+1}, \action_{t})$. 
Each entry in $\mathbf{k}$ can be interpreted as the relative stiffness of the spring associated with a particular constraint.

The constraints we use are: $\mathbf{C}(\statesim_{t+1}, \action_{t}) = \mathbf{C}_{m}(\statesim_{t+1}) \cup \mathbf{C}_{a}(\statesim_{t+1}, \action_{t}) \cup \mathbf{C}_{b}(\statesim_{t+1})$.
The mesh deformation constraints, $\mathbf{C}_{m}$, include distance constraints defined on every mesh edge and bending constraints between neighboring mesh faces. 
We extend the XPBD framework by incorporating two new geometric constraints to enable robot end-effector control and environment attachment points. 
These robot grasp-related constraints are formulated as $\mathbf{C}_{a}$,  and $\mathbf{C}_{b}$  describes attachment-related constraints. 
These new constraints are discussed below.

We solve the minimization of $\energy(\statesim_{t+1}, \action_{t})$ in Eq.~\ref{eqn:energy_minimization_xpbd} by iteratively updating particle positions, initialized at $\statesim_{t}$, to satisfy the constraints using the following update rule: 
\begin{equation}\label{eqn:iterative_update}
    \begin{split}
        \Delta \mathbf{x} &= \mathbf{M}^{-1} \nabla \mathbf{C}^\top \Delta \lambda\\
        \Delta \mathbf{\lambda} &= - \Bigl( \nabla \mathbf{C} \mathbf{M}^{-1} \nabla^{\top} \mathbf{C} + \mathbf{\tilde \alpha}\Bigr)^{-1} \Bigl(\mathbf{C} + \mathbf{\tilde \alpha} \mathbf{\lambda} \Bigr) \\
        \mathbf{\tilde \alpha} &= \mathtt{diag}(\mathbf{k})^{-1} / \Delta t^2
    \end{split}
\end{equation}
in which $\mathbf{M}$ is a diagonal mass matrix, and $\lambda$ is a Lagrange multiplier. For more details on this update rule, see~\cite{macklin2016xpbd}.

We summarize the XPBD simulator as follows:
\begin{equation}\label{eqn:pbd_sim}
\begin{split}
    \mathbf{x}_{t+1} & = f(\mathbf{x}_{t}, u_{t}, \mathbf{b}), \\ 
    s.t., \mathbf{C}_m(\mathbf{x}_{t+1})&=\mathbf{0},\mathbf{C}_a(\mathbf{x}_{t+1}, u_t) = \mathbf{0},\mathbf{C}_b(\mathbf{x}_{t+1})=\mathbf{0}
\end{split}
\end{equation}
where $\bel \subset \mathbf{k}$ are constraint strength parameters for the $\mathbf{C}_{b}$ that we will elaborate on below.

\textbf{End-effector Grasp Constraints $\mathbf{C}_a$.} 
To enable control, 
the tissue mesh must be constrained by the current grasp point of the robot. 
Note that at time $t-1$, the robot selects as its action $u_{t-1}$ the next location of the end effector. 
Therefore, the current grasp location at time $t$ is $u_{t-1}$. 
This grasping point is treated as an infinite-mass virtual particle that is connected to points on the mesh within a local neighborhood $\mathbf{x}_t^g \subset \mathbf{x}_t$.
The $i$-th particle is included in $\mathbf{x}^g$ if $\|x^i_1 - u_0\|_2$, \textcolor{rebuttal}{the initial distance between the $i^{th}$ particle and chosen grasp point at time $t=1$}, is less than radius $r$. 
The robot can grasp different parts of the mesh by choosing a different initial grasp point $u_0$.
In our implementation, $\mathbf{x}^g$ is constrained by linear springs connecting each entry of $\mathbf{x}^g$ s.t. $u_t$:  $C^j_a(x_{t+1}, u_t) = \|x^j_{t+1} - u_t\|_2 - \|x^j_1 - u_0\|_2, x^j \in \mathbf{x}^g, C^j_a \in \mathbf{C}_a$. 
We assume knowledge of the spring constants $\mathbf{k}_{a}\subset\mathbf{k}$ 
associated with $\mathbf{C}_{a}$.

\begin{figure}[t]
    \centering
    \includegraphics[width=1\linewidth]{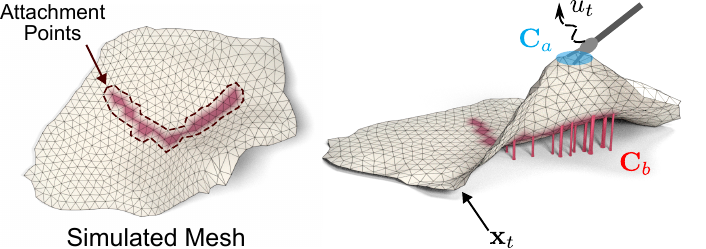} 
    \caption{An illustration of the spring boundary and grasping constraints discussed in section \ref{sec:PBD_simulator}. 
    Attachment points on the tissue are depicted in red and are modeled using springs of varying stiffness that connect the tissue to the environment. 
    Meanwhile, the grasp is modeled as an infinite-mass virtual particle that is connected via springs to a small neighborhood of particles highlighted in blue.
    }
    \label{fig:spring_boundary}
    \vspace{-0.5em}
\end{figure}

\textbf{Spring Boundary Constraints $\mathbf{C}_b$.}
    We formulate attachments between a tissue and its surrounding environment with spring boundary constraints similar to \cite{xu2022identification}. 
    The idea, illustrated in Fig.~\ref{fig:spring_boundary}, assumes that a zero resting-length spring with unknown stiffness exists between every simulation particle $x^i_t$ and its corresponding infinite-mass \textit{virtual particle} whose position is fixed at $x_0^i$. 
    Each attachment can be expressed with a geometric constraint $C_b^i(\mathbf{x}) = || x^i_t - x^i_0 ||_2, x^i \in \mathbf{x}$, and the spring constant or the strength of the corresponding spring boundary constraint on particle $x^{i}$, \textcolor{rebuttal}{denoted with $b^i$}. 
    $\bel = [b^{1} \dots b^{n}]^{\top} \subset \mathbf{k}$ denotes all spring constants between the particles and the environment. 
    Particles attached to the environment have a high stiffness, $b^{i}>0$. Particles in free regions have $b^{i} = 0$. 
    
We aim to estimate unknown spring boundary parameters, $\bel^{*}_{t}$,
from observations of mesh deformations.
The proposed simulation is implemented in PyTorch and is fully differentiable using PyTorch's automatic differentiation package.

\subsection{EKF-Based Boundary Estimation}\label{sec:estimation_framework}
    
    We propose a probabilistic estimation framework to track distributions, $\bel_{t}$, over true boundary parameters $\bel^{*}_{t}$, in a changing environment. 
    We represent the distribution over our boundary parameters using a multivariate Gaussian distribution. 
    This enables efficient tracking of our boundary estimates using a Kalman filter-based approach. 
    The differentiable XPBD simulation allows us to locally linearize the tissue model to fit an extended Kalman filter (EKF) formulation. 
    The approach enables us to treat the mesh particle states as indirect observations of the boundary parameters, which are our latent random variable states.  
    
    \textbf{Motion Model.} At any given timestep $t$, our belief of the current boundary parameters is represented by the random variable $\bel_{t} \sim \mathcal{N}(\mub_{t}, \Sigmab_{t}) $.
    The motion model describes the change in the underlying boundary between timesteps.
    This captures topology-changing actions that directly modify the boundary parameters.
    We formulate a boundary-changing action as $\delta \bel_{t} \in \mathbb{R}^{n}$.
    We assume such changes can only be the result of special actions such as cutting and suturing and not of robot grasping. 
    As a result, we control the value of $\delta \bel_{t}$. 
    For later convenience, we denote the last topology-changing action's time step as $t_c$. 
    Our motion model is defined as: 
    \begin{equation} \label{eqn:motion_model}
    \begin{split}
        \bel_{t+1} & = \motionmodel(\bel_{t}, \delta \bel_{t}, \motionnoise_{t}) \\
        & = \bel_{t} + \delta \bel_{t} + \motionnoise_{t} \\
        \motionnoise_t & \sim \mathcal{N}(0, \motioncov_{t})
    \end{split}
    \end{equation}
    In the absence of topological changes, $\delta \bel_{t} = 0$. 
    During active changes, elements in $\delta \bel_{t}$ will be assigned non-zero values, especially in regions closer to the topological change. 
    For example, $\delta \bel_{t}$ will be given negative values to indicate cutting or positive values to indicate the addition of attachment points through actions like suturing.
    $\motionnoise_{t}$ is an additive Gaussian motion model noise. 
    This noise has covariance $\motioncov_{t}$ with increased variance near the modified region, indicating uncertainty about the outcomes of the topology-changing event. 

    \textbf{Observation Model.} Given that the boundary parameters are not directly observable, we observe them indirectly through the state of the tissue, $\stateref$, which depends on $\bel_t$. 
    The observation model $\obsmodel$ is defined as:
    \begin{equation}\label{eqn:ekf_obs_model}
    \begin{split}
        \statesim_{t+1}     & = \obsmodel(\bel_{t+1}, \stateref, \action_{t}, \obsnoise_{t}) \\
                        & = \pbd(\stateref, \action_{t}, \bel_{t+1}) + \obsnoise_{t+1} \\
        \obsnoise_{t+1}   & \sim \mathcal{N}(0, \obscov_{t+1})
    \end{split}
    \end{equation}
    Here, we initialize our XPBD simulation, $\pbd$, using the last observed tissue surface $\stateref$ and simulate one time step forward with the updated belief of the boundary parameters $\bel_{t+1}$, updated using the motion model $\motionmodel$, and the control action $\action_{t}$. 
    The covariance matrix $\obscov_{t+1}$ directly characterizes the noise in observations of the tissue surface. 
    We use these elements to establish the following EKF equations to track the changes in our boundary parameter belief, represented by $\mub_{t}$ and $\Sigmab_{t}$.

    
    \textbf{Prediction :}
    \begin{equation} \label{eqn:predict_step}
        \begin{split}
            \mub_{t+1|t} = \mub_{t|t} + \delta \bel_{t} \\
            \Sigmab_{t+1|t} = \Sigmab_{t|t} + \motioncov_{t}
        \end{split}
    \end{equation}
    \textbf{~~~Update :}
    \begin{equation} \label{eqn:update_step}
        \begin{split}
            \mub_{t+1|t+1} &= \mub_{t+1|t} + K_{t+1} \mathbf{\tilde{y}}_{t+1}\\
            \Sigmab_{t+1|t+1} & = (I - K_{t+1}\beljacob_{t+1})\Sigmab_{t+1|t}
        \end{split}
    \end{equation}
    where the observation residual $\mathbf{\tilde{y}}_{t+1}$, and the Kalman gain $\kalmangain_{t+1}$ are given by:
    \begin{equation} \label{eqn:kalman_gain}
        \begin{split}
        \mathbf{\tilde{y}}_{t+1} & = \staterefp - h(\mub_{t+1|t}, \stateref, \action_{t}, \obsnoise_{t}) \\
        \kalmangain_{t+1} & = \Sigmab_{t+1|t}\beljacob_{t+1}^{\top} \residualcovariance_{t+1}^{-1}
        \end{split}
    \end{equation}
    with the residual covariance $\residualcovariance_{t+1}$, the observation Jacobians $\beljacob_{t+1}$, and covariance Jacobians $\noisejacob_{t+1}$ as:
    \begin{equation} \label{eqn:kalman_jacobians}
        \begin{split}
        \residualcovariance_{t+1} & = \beljacob_{t+1}\Sigmab_{t+1|t}\beljacob^{\top}_{t+1} + \noisejacob_{t+1}\obscov_{t+1}\noisejacob_{t+1}^{\top}\\
        \noisejacob_{t+1} &= \frac{\partial \obsmodel}{\partial \obsnoise}\bigg|_{\mub_{t+1|t}} = I \\
        \beljacob_{t+1} & = \frac{\partial \obsmodel}{\partial \bel} \bigg|_{\mub_{t+1|t}} = \frac{\partial \pbd(\stateref, \action_{t}, \bel_{t+1})}{\partial \bel} \bigg|_{\mub_{t+1|t}}
        \end{split}
    \end{equation}
    Here, $\pbd(\stateref, \mub_{t+1|t}, \action_{t})$ outputs the expected tissue surface after using the XPBD simulator to forward simulate one timestep with action $\action_{t}$, using the mean of the current estimated boundary belief, $\mub_{t+1|t}$.
    \textbf{Multiple Shooting Metric. } 
    To enhance the robustness of the estimation, we modify the EKF to use previous observations \textcolor{rebuttal}{within a specified horizon}. 
    \textcolor{rebuttal}{This is possible with the assumption that no topology changing events occur within the considered observation window and that boundary parameters are constant unless changed by these controlled topology changing events.}
    We use the multiple shooting method to encode previous observations \cite{heiden2022probabilistic}; this approach samples from previously observed trajectories,
    forward simulates from those samples using the current belief 
    and penalizes the deviation of predicted particle states from the reference observations. 
    For simplicity, we set the prediction horizon to 1. The new EKF update step is defined as 
    \begin{equation} \label{eqn:update_step_multiple_shooting}
        \begin{split}
            \mathcal{X}^{\text{ref}}_{M+1} = &\begin{bmatrix}\mathbf{x}^{\text{ref}}_{m1+1}\\\vdots\\\mathbf{x}^{\text{ref}}_{mk+1}\end{bmatrix},\ 
            \mathcal{X}^{\text{pred}}_{M+1} = \begin{bmatrix}\pbd(\mathbf{x}^\text{ref}_{m1}, \action_{m1}, \bel_{t+1|t})\\\vdots\\\pbd(\mathbf{x}^\text{ref}_{mk}, \action_{mk}, \bel_{t+1|t})\end{bmatrix}\\
             J_{M+1} = &\begin{bmatrix}\displaystyle\frac{\partial \pbd(\mathbf{x}^\text{ref}_{m1}, \action_{m1}, \bel_{t+1|t})}{\partial \bel} \bigg|_{\mub_{t+1|t}}\\\vdots\\\displaystyle\frac{\partial \pbd(\mathbf{x}^\text{ref}_{mk}, \action_{mk}, \bel_{t+1|t})}{\partial \bel} \bigg|_{\mub_{t+1|t}}\end{bmatrix}, R_{M+1} = \begin{bmatrix}I\\\vdots\\I\end{bmatrix}\\
            \mub_{t+1|t+1} =& \mub_{t+1|t} + K_{t+1|t} \bigl( \mathcal{X}^{\text{ref}}_{M+1} - \mathcal{X}^{\text{pred}}_{M+1}\bigr) \\
            \Sigmab_{t+1|t+1} = &(I - K_{t+1|t}\beljacob_{M+1})\Sigmab_{t+1|t}\\
             K_{t+1|t} =& \Sigmab_{t+1|t}\beljacob_{M+1}^{\top}\big(\beljacob_{M+1}\Sigmab_{t+1|t}\beljacob^{\top}_{M+1} + \\
              &\noisejacob_{M+1}\obscov_{t+1}\noisejacob_{M+1}^{\top}\big)^{-1}
        \end{split}
    \end{equation}
    Here, $M=\{m_1,...,m_k\}$ are $k$ samples drawn by first selecting $m_k=t$ and uniformly sampling $k-1$ in the range $[t_c, t-1]$. 
    $t_c$ is the time step of the last topology-changing event. 
    We do not include time steps before this as the previous topology has been altered and is therefore irrelevant.

\subsection{Joint Boundary Parameter and State Estimation for Real-world Scenarios}\label{sec:joint_estimation}
 Our estimation framework requires the knowledge of reference particle positions $\{\mathbf{x}^\text{ref}_t\}$ to update spring boundary parameters. 
 This information is easily accessible in simulation but requires additional effort in the real world. 
Let $\mathcal{T}^{\text{ref}}$ denote the real tissue. 
Via robot perception, we can access dense surface point clouds $Q_t$ of the tissue. 
$\mathbf{x}^\text{ref}_t$ can be estimated by finding a configuration of mesh particles that matches $Q_t$ and respects geometric constraints.
   
   Similar to \cite{liang2023realtosim}, we achieve this by introducing an observation matching condition into the XPBD's simulation's iterative constraint solving procedure.
   Specifically, the first row of Eq.~\ref{eqn:iterative_update} is modified to
   \begin{equation}\label{eqn:pbd_plus_residual}
        \begin{split}
            \Delta \mathbf{x} &= \mathbf{M}^{-1} \nabla \mathbf{C}^\top \Delta \lambda - \alpha \nabla_\mathbf{x} \mathcal{P}({}^i\mathbf{x}, Q_t)\\
        \end{split}
    \end{equation}
    where ${}^i\mathbf{x}$ represents intermediate simulation particles in the iterative process. 
    Additionally, $\alpha$ is a step size, and $\mathcal{P}(\cdot)$ is a loss metric that measures the difference between $^i\mathbf{x}$ and $Q_t$. 
    To compute the metric, another dense point cloud $P_t$ is sampled from the simulation mesh using differentiable point cloud sampling \cite{sundaresan2022diffcloud}. The loss metric is then defined by the chamfer distance between two point clouds
    \begin{equation*}
        \begin{split}
            \mathcal{P}({}^i\mathbf{x}, Q_t) &= \mathcal{CD}({}^iP, Q_t) \\
            &= \sum_{p\in {}^iP} \min_{q\in Q_t}\|p - q\|^2_2  + \sum_{q\in Q_t} \min_{p\in {}^iP}\|q -  p\|^2_2.\\
        \end{split}\label{eqn:chamfer}
    \end{equation*}

    Note that in Eq.~\ref{eqn:pbd_plus_residual} for estimating $\stateref$, $\Delta \lambda$ depends on the current parameter estimate $ \mub_{t}$.
    Therefore, the two variables, $\stateref$ and $\mub_{t}$, are coupled through the registration of the point cloud data. This motivates us to combine the estimation of $\stateref$ and the update step of EKF into one joint estimation problem, where both sub-problems are iteratively estimated:
    \begin{equation}
        \begin{split}
            \mub_{t+1}, \Sigmab_{t+1} & = \text{EKF\;Update}(\mub_{t}, \Sigmab_{t}, \{\mathbf{x}^\text{ref}_{t'}\}_{t'=t_c}^{t}, M)\\
            \mathrm{s.t.} \ \ \mathbf{x}^\text{ref}_{m+1} & = f'(\mathbf{x}^\text{ref}_m, u_m, \mub_t, \pcl_m),\ m\in M \\
             \mathbf{C}(\mathbf{x}^\text{ref}_{m+1})&=\mathbf{0},\ \mathbf{C}_a(\mathbf{x}^\text{ref}_{m+1}, u_m),\ \mathbf{C}_b(\mathbf{x}^{\text{ref}}_{m+1})=\mathbf{0}
        \end{split}
    \end{equation}
    where $f'(\cdot)$ is a modified XPBD equation that uses Eq. ~\ref{eqn:pbd_plus_residual}.

\subsection{\textcolor{rebuttal}{Objectives and Constraints for Active Sensing Control}}\label{sec:active_sensing}
    This section delves into active sensing methodologies that we build on top of our estimation framework. 
    We leverage our current belief of the tissue boundary parameters to inform the selection of next step actions and encourage the discovery of additional tissue attachment points.  
    We investigate four unique active control objectives. 
    These objectives are \textbf{Boundary Entropy Minimization}, \textbf{Uncertainty-Weighted Displacement (UWD)}, \textbf{Boundary Energy Minimization}, and \textbf{In-Workspace Constraints}. 
    Each objective offers a unique perspective on guiding actions for exploration. 
    
    \textbf{Boundary Entropy Minimization.}
    Entropy is a measure of uncertainty of a probability distribution, where a higher entropy indicates higher uncertainty. 
    We therefore seek to take actions that reduce the entropy $\entropy$ of our belief distribution $\bel_{t}$: 
    \begin{equation}\label{eqn:gaussian_entropy}
    \begin{split}
        \entropy(\mub_{t+1}) & = - \int p(\mub_{t+1})\ln p(\mub_{t+1}) {\rm d} \bel
    \end{split}
    \end{equation}
    As a result, we represent the entropy's evolution as a function of the estimated boundary $\bel_{t+1}$ at time $t+1$ by employing the update step equations from the EKF as outlined in Eq. \ref{eqn:update_step}.
    \begin{align}
            \entropy(\mub_{t+1}) 
            & = \frac{D}{2}(1 + \ln(2 \pi)) + \ln\left(|\Sigmab_{t+1}| \right) \label{eqn:kalman_entropy}\\
            & \approxeq \ln\left(|\Sigmab_{t+1}| \right)\label{eqn:kalman_entropy_approx}
    \end{align}
    where $D$ is the dimensionality of the multivariate Gaussian distribution.
    As the first term in Eq.~\ref{eqn:kalman_entropy} is a constant $c=\frac{D}{2}(1 + \ln(2 \pi))$,  it can be disregarded during optimization.
    We employ this objective to select the control action that minimizes the entropy through optimization in the following section.


\textbf{Uncertainty-Weighted Displacement (UWD) Maximization.}
    \input{main_tex_files/methods_active_sensing_displacement}
\textbf{Boundary Constraint Energy Minimization.}
    During active exploration, we want to pose safety constraints that avoid exerting excessive force at connected regions. 
    In surgery, this is critical to prevent unwanted tearing and bleeding.
    We formulate an energy-based safety metric to penalize high elastic potential energy at the boundary. 
    The elastic potential energy of boundary constraints is computed similarly to Eq.~\ref{eqn:elastic_potential}:
    \begin{equation}\label{eqn:boundary_elastic_potential}
    \begin{split}
                \energy_b \left(\stateref, \action_{t}, \mub_{t} \right) & = \frac{1}{2}\ \mathbf{C}_b(\statesim_{t+1})^{\top} \mathtt{diag}(\mub_t) \mathbf{C}_b(\statesim_{t+1}) \\
    \end{split}
    \end{equation}

\textbf{In-Workspace Constraint.}
    In surgery, the workspace of the robot is often constrained due to the ports used to insert the laparoscopic tools and obstructions created by organs and other tissues. 
    We formulate a position-based workspace constraint to limit the range of the next grasp point.
    This penalizes actions that exit a constraint region. 
    \textcolor{rebuttal}{For example, in our work we consider the constraint where the robot should not pull the tissue downward through the ground plane, which represents the anchoring tissue}
    We represent this constraint function as: 
    \begin{equation} \label{eqn:in_workspace}
        \begin{split}
            \inworkspace(\action_{t}) = e^{-\min([\action_{t}]_z, 0)}
        \end{split}
    \end{equation}
    where $[\action_{t}]_z$ is the height of the grasp point from action $\action_{t}$. 
     \textcolor{rebuttal}{While in surgery the environment can also deform, here we assume the ground plane doesn't change.}
     \textcolor{rebuttal}{Though we use the constraint in Eq. \ref{eqn:in_workspace}, this constraint can be tailored to restrict the choice of actions for the scenario at hand.} 

\subsection{Optimization-based Controller Design} \label{sec:active_controller_design}

Based on our choices of active sensing objectives, we present two variants of controllers that both take advantage of the differentiability of the PBD simulator. 
They are 1) a local gradient-based action controller 
and 
2) a sampling-based large-step action controller.

\textbf{Local Gradient-based Action Control.}
\label{sec:local_feedback_control}We 
locally solve for the action that minimizes the entropy of our belief distribution on the boundary parameters without damaging the tissue at the boundary or violating in-workspace constraints: 
    \begin{equation} \label{eqn:local_minimization_entropy}
        \begin{split}
            \arg\min_{\action_{t}} ~ & \entropy + \boundaryenergy \\
            \mathrm{s.t.} ~~~  & \inworkspace(\action_{t}) \geq \inworkspace_{\min}
        \end{split}
    \end{equation}
    

Nevertheless, dealing with this high-dimensional optimization challenges traditional solvers. 
To mitigate this, we relaxed the constraints using a penalty-based method, addressing Eq. \ref{eqn:local_minimization_entropy} as an unconstrained optimization problem, solved using gradient descent. 
This approach capitalizes on the differentiability of our XPBD simulator.
\begin{equation} \label{eqn:local_entropy_loss}
    \begin{split}
        \min_{\action_{t}} \textcolor{rebuttal}{\mathcal{L_H}} & = \min_{\action_{t}} \entropy + \boundaryenergy + \sigma \norm{\max\big\{ 0,  \inworkspace_{\min} - \inworkspace \big\} }\\
    \end{split}
\end{equation}
where $\sigma$ denotes a weight.
Gradient descent iteratively updates the best local action to take: 
    \begin{equation} \label{eqn:local_action_update_entropy}
        \begin{split}
            \action_{t} = \action_{t-1} - \alpha \nabla_{\action_{t}} \mathcal{L}
        \end{split}
    \end{equation}
\textcolor{rebuttal}{
To simplify the gradient calculation, we replace entropy ($\mathcal{H}$) in Eq.~\ref{eqn:local_entropy_loss} with uncertainty-weighted displacement (UWD), $\mathcal{D}$, from Eq.~\ref{eqn:uncertain_weighted_displacement}.
This substitution is made possible given the proposition \ref{appendix:relationship_displacement_entropy}, in the appendix, which illustrates the relationship between entropy, $\mathcal{H}$, and UWD, $\mathcal{D}$. 
With this we have the new loss function: 
}
\begin{equation} 
\label{eqn:local_loss_displacement}
    \begin{split}
        \textcolor{rebuttal}{\mathcal{L_D}} = -\weighteddeform + \boundaryenergy + \sigma \norm{\max\big\{ 0,  \inworkspace_{\min} - \inworkspace \big\} } \\
    \end{split}
\end{equation}

Once optimized, the robot executes action $\action_{t}$, obtains an observation $\staterefp$, updates the belief to be $\mub_{t+1}$, and then repeats this process. 

In practice, we noticed that this local approach stopped deforming the tissue after a few iterations. 
This is attributed to vanishing gradients, signifying that the optimization has reached a point where local-step control actions have a negligible impact on reducing the loss term. 

\textbf{Sampling-based Large Step Action Control.}
    To address the challenges posed by local minima and gradient vanishing issues, we explore longer-horizon action sequences for more information acquisition. 
    Although trajectory optimization can mitigate these issues, it becomes computationally demanding, especially for high degree-of-freedom deformable objects. 
    Therefore, we introduce a sampling-based large step controller to overcome the limitations associated with gradient locality.

    Instead of considering a long horizon with multiple control steps, we optimize a set of large-step control samples $\mathbb{S}$ from the workspace in parallel. 
    For each sample, we take a single large control step from the current grasp point. 
    We reformulate the following minimization problem:
    \begin{equation}\label{eqn:large_step_minimization}
       \begin{split}
        s_{k}^{*} & = \arg\min_{s_{k} \in \mathbb{S}} \textcolor{rebuttal}{\mathcal{L_D}}(\mathbb{S}) \\
        \mathrm{s.t.}~~~~ & \mathbb{S} = \big\{s_1, s_1, \cdots, s_k\big\} \\ 
                          & s_{k} = s_{k}  - \alpha \nabla_{\action} \textcolor{rebuttal}{\mathcal{L_D}} \\
       \end{split} 
    \end{equation}  
    After finding the best large step action $s^*$ from the set $\mathbb{S}$, we move the current grasp point $\action_{t-1}$ towards it using truncated control length $\gamma$.
    \begin{equation}\label{eqn:large_step_action}
        \begin{split}
            \action_{t}^{*} = \action_{t-1} + \gamma \frac{s_{k}^{*}-\action_{t-1}}{ \norm{s_{k}^{*}-\action_{t-1}} }
        \end{split}
    \end{equation}
    \textcolor{rebuttal}{We re-sample the large-step control samples $\mathbb{S}$ between each action by uniformly sampling the workspace. 
    To encourage smooth trajectories and a balance between exploration and exploitation, we retain the top $10\%$ of samples in $\mathbb{S}$ (samples with the lowest loss values, $\mathcal{L}_{\mathcal{D}}$) from the previous iteration, when re-sampling $\mathbb{S}$.}
    It is worth noting that the optimization of sample sets can be computed concurrently on the GPU for accelerated processing.
    
    
    Our full algorithm is summarized in Alg. \ref{alg:full_algorithm}. 
    In line 4-12, we handle topological changes using the prediction step of our probabilistic estimation framework (section \ref{sec:estimation_framework}). 
    Following that, visual feedback is used to update our boundary parameter belief in a joint estimation manner in line 13-20 (section \ref{sec:estimation_framework}, \ref{sec:joint_estimation}). Finally, the proposed active sensing objectives and controllers section \ref{sec:active_sensing}, \ref{sec:active_controller_design} can be used to determine the next action in line 21-25.

%% file: main_tex_files/methods_active_sensing_displacement.tex
The strategy of minimizing entropy is heavily dependent on the accuracy of the current boundary parameter estimate. In a complex deformable environment, the greedy minimization of this strategy with imperfect boundary estimates is prone to local minima. Moreover, the computational complexity associated with entropy propagation, i.e., $\nabla_{\action_{t}}{\Sigmab_{t+1|t+1}}$, limits the applicability of this heuristic in sampling-based and multi-step trajectory optimization approaches.

By introducing a heuristic function for uncertainty-weighted displacement (UWD), denoted as $\weighteddeform$, our new metric relies exclusively on the covariance matrix $\Sigmab_{t}$ updated from the preceding step. 
Notably, this matrix is independent of the current step motion, as evidenced by $\nabla_{\action_{t}}{\Sigmab_{t}} = 0$.
To address challenges in the previous metric, particularly in computing intense gradients, our proposed function directly calculates deformation displacement. 
The objective is expressed as:
\begin{equation}\label{eqn:uncertain_weighted_displacement}
    \begin{split}
        \weighteddeform(\stateref, \action_{t}, \mub_{t}) & = \norm{ \wideparen{\Delta\statesim_{t+1}} \cdot \Sigmab_{t}}
    \end{split}
\end{equation}
\begin{equation}\label{eqn:uncertain_weighted_displacement_values}
    \begin{split}
        \statesim_{t+1}& = \pbd\left(\stateref, \action_{t}, \mub_{t+1|t}  \right) \\
        \Delta \statesim_{t+1} & = \statesim_{t+1} - \statesim_{0} \\
        \wideparen{\Delta\statesim_{t+1}} & = \sum_i^n \mathbf{e}_i \mathbf{e}^{\top}_i \otimes [\Delta \statesim_{t+1}]_i. \\
    \end{split}
\end{equation}
Here, $\Delta \statesim_{t+1}$ represents the simulated displacement of the tissue mesh particles at time $t+1$ from their original positions at $t=0$, resulting from the forward XPBD simulation $\pbd\left(\stateref, \action_{t}, \mub_{t+1|t}  \right)$ under the action $\action_{t}$, and the current updated boundary belief $\mub_{t+1|t}$. 
A concatenated-matrix form $\wideparen{\Delta\statesim_{t+1}}$ of this displacement is computed by
letting $\mathbf{e}_i$ be the $i$-th column vector of the standard basis of $\mathbb{R}^n$, and $[\Delta \statesim_{t+1}]_i \in \mathbb{R}^{3 \times 1}$ 
be the deformation displacement of the $i$-th particle in $\statesim_{t+1}$. 
Then, a deformation displacement matrix $\wideparen{\Delta\statesim_{t+1}}$ can be derived with $\otimes$ denoted as Kronecker product operation, facilitating the proper weighting of particle displacement by the covariance of our current boundary belief $\Sigmab_{t}$.

The objective of maximizing the $\weighteddeform$ in Eq.~\ref{eqn:uncertain_weighted_displacement} is akin to maximizing tissue displacement, with each region weighted by the uncertainty of the current belief in that location. 
This heuristic consequently emphasizes the enhancement of displacement in regions of the tissue surface characterized by higher uncertainty. 
In the Appendix \ref{appendix:relationship_displacement_entropy}, we have presented a proposition along with an explanation demonstrating that this objective can still guide efficient entropy minimization without directly relying on the future prediction of the covariance matrix, i.e., $\Sigmab_{t+1|t+1}$.

%% file: main_tex_files/evaluations.tex
\begin{figure}[!t]
    \centering
    \includegraphics[width=1\linewidth]{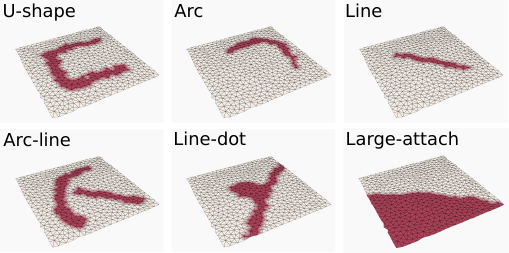} \vspace{-1em}
    \caption{
    The images show the ground truth attachment points in red using spring boundary constraints for the simulation test cases in our experiments. 
    }
    \label{fig:simulation_testcases}\vspace{-1em}
\end{figure}

In our experiments, we seek to evaluate our framework's capabilities in 1) estimation accuracy in both simulated and real scenarios, 2) actively maximizing information gain while considering safety, and 3) intelligently adapting to topological changes. 
The full capabilities of our approach are evaluated in simulated environments in Sections \ref{sec:simulation_estimation_accuracy}, \ref{sec:active_exploration_experiment}, and \ref{sec:reestimation_experiment}.
In Section \ref{sec:real_world_estimation_accuracy}, we analyze \textcolor{rebuttal}{only} our \textcolor{rebuttal}{estimation} performance in real-world tissue scenarios.


\subsection{Evaluation Metrics}
 \textbf{Accuracy.}
 Estimation accuracy is measured in two terms. They are 1) the percentage of correct detection (PCD): the number of correct attachment point detections over the total number of detections, and 2) the percentage of uncovered ground truth (PUG): the number of uncovered attachment points divided by the total number of connections. 
 The $i$th mesh point $\statesim^{i}_t$ is considered attached at time $t$ if $b^i_t > b_{thresh}$. 
 In our experiments $b_{thresh} = 0.05$. 
 
 Due to observational noise and limitations posed by discretization, estimated attachment points can be spatially close to ground truth but not perfectly overlap. 
 Therefore, we report results from performing a dilation operation to expand the ground truth attachment region spatially and estimated connections for computing PCD and PUG, respectively. 
 Dilation is implemented with graph convolution on simulation meshes.

\textbf{Active Sensing Performance.} 
Different active sensing methods are compared in terms of the amount of entropy reduction within a fixed number of timesteps. 
All comparison active sensing methods are built on top of the proposed EKF-based estimation framework.
Their entropy is calculated using Eq. \ref{eqn:gaussian_entropy}. 
We also evaluate the safety performance of comparison methods in terms of the mean boundary energy calculated using ground truth boundary stiffness and Eq. \ref{eqn:boundary_elastic_potential}. In this term, a larger value means a higher chance of unsafe tearing.

\begin{table}[t]
\setlength\tabcolsep{1.5em}
\centering
\caption{Comparison of boundary estimation accuracy after 4 grasp sequences. Results are compared between our methods and the Adam optimizer. The proposed outperforms better than Adam in all test cases.
}
\begin{adjustbox}{width=0.45\textwidth}
    \begin{tabular}{l|cc|cc}
    \toprule
    & \multicolumn{2}{|c}{PCD \upgreenarrow}& \multicolumn{2}{|c}{PUG \upgreenarrow}\\ \midrule
 Cases & Adam & Ours & Adam & Ours\\ 
  \midrule
Arc & 82.1  & 100 & 59.3 & 87.1\\
Line & 85.7  & 100 & 40.0 & 43.3\\
Line-dot & 68.0  & 96.0 & 34.7 & 48.9\\
Arc-line & 83.8  & 97.3 & 51.8 & 65.5\\
U-shape & 87.3 & 97.5  & 52.9 & 76.0 \\
    \bottomrule
\end{tabular}
\end{adjustbox}
\label{tbl:accuracy_simulation}
\end{table}

\subsection{Estimation Capabilities in Simulation}\label{sec:simulation_estimation_accuracy}
We first evaluate the estimation accuracy of the proposed method in a simulated environment with a pre-defined set of robot actions (i.e. no active sensing). 
We manually define different shapes of connected regions that are visualized in Fig.~\ref{fig:simulation_testcases}.
The boundary parameters at the attached regions are set with a ground truth stiffness $\bel^*=0.1$. 
We design control sequences that can potentially reveal the entire shape of the attached region.
For each test case (case \textit{Large-attach} excluded in this experiment), four corners of the squared mesh are sequentially grasped and lifted up to a height that is roughly equal to half of the mesh width.
In each test case, we initialize our belief on the boundary parameters with the assumption that there are no attachment points by setting the mean of our boundary belief to be very small at $\mub_{0}=1e-4$. 
For our baseline, we use a stochastic gradient descent (Adam) algorithm that optimizes a similar multiple shooting metric to directly update a deterministic estimate of the boundary parameters by differentiating through the XPBD simulator. 
\textcolor{rebuttal}{To update the boundary parameters using Adam, we first compute a multiple shooting based loss: 
$$\mathcal{L}_{Adam} = \frac{1}{k}|| \mathcal{X}^{\text{ref}}_{M+1} - \mathcal{X}^{\text{pred}}_{M+1} ||^{2}_{2}$$
Where $\mathcal{X}^{\text{ref}}_{M+1}$ and $\mathcal{X}^{\text{pred}}_{M+1}$, as defined in Eq. \ref{eqn:update_step_multiple_shooting},  consist of $k$ ground truth observations and predicted samples respectively. 
Leveraging the differentiability of our XPBD simulation, we compute a backward pass to get the gradients  with respect to the boundary parameters $\mub_{t}$ that minimize this loss. 
We use these gradients with pytorch's Adam optimizer to directly step and update the estimated boundary parameters $\mub_{t}$. 
}

\begin{figure}[t]
    \centering
    \includegraphics[width=1\linewidth]{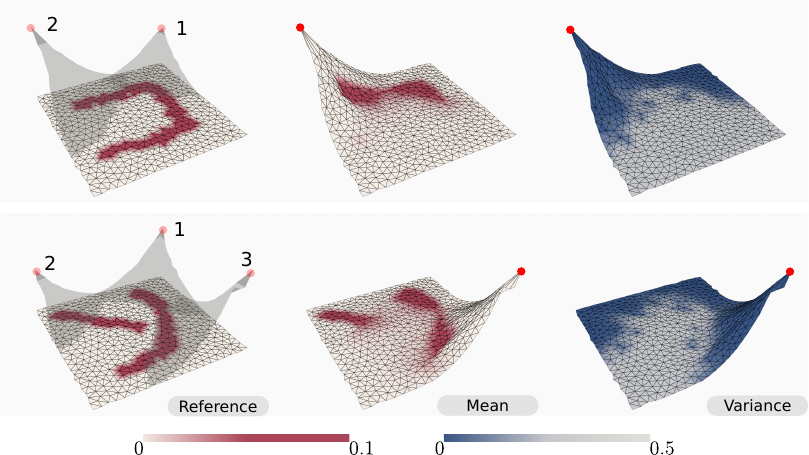} \vspace{-1.3em}
    \caption{Example results from our estimation framework on simulated environments where the order of the grasps is numbered, and the boundary is highlighted in red on the left-most column.
    The final result from our proposed method is shown in red in the middle column.
    Finally, the confidence (inverse of variance) of our estimation is shown in blue in the rightmost column.
    We can see how the variance has decreased in the regions where the trajectories have displaced the tissue from its original state, and the mean estimate has converged close to the reference values.
   }
\label{fig:simulation_estimation}\vspace{-1.5em}
\end{figure}

The complete set of results after the execution of the entire trajectory is shown in Table~\ref{tbl:accuracy_simulation}. Our proposed approach performs better than the baseline, uncovering more attachment regions with improved accuracy under the same manipulation sequences. 
Example results from our estimation method are visualized in Fig.~\ref{fig:simulation_estimation}, which demonstrates that the proposed EKF estimator is able to identify partial attachment regions, as shown by the current mean of the boundary parameter estimate. 
Regions that have not been influenced by previous control sequences remain uncertain, as shown by the variance.


\begin{figure*}[t]
    \centering
    \includegraphics[width=0.99\linewidth]{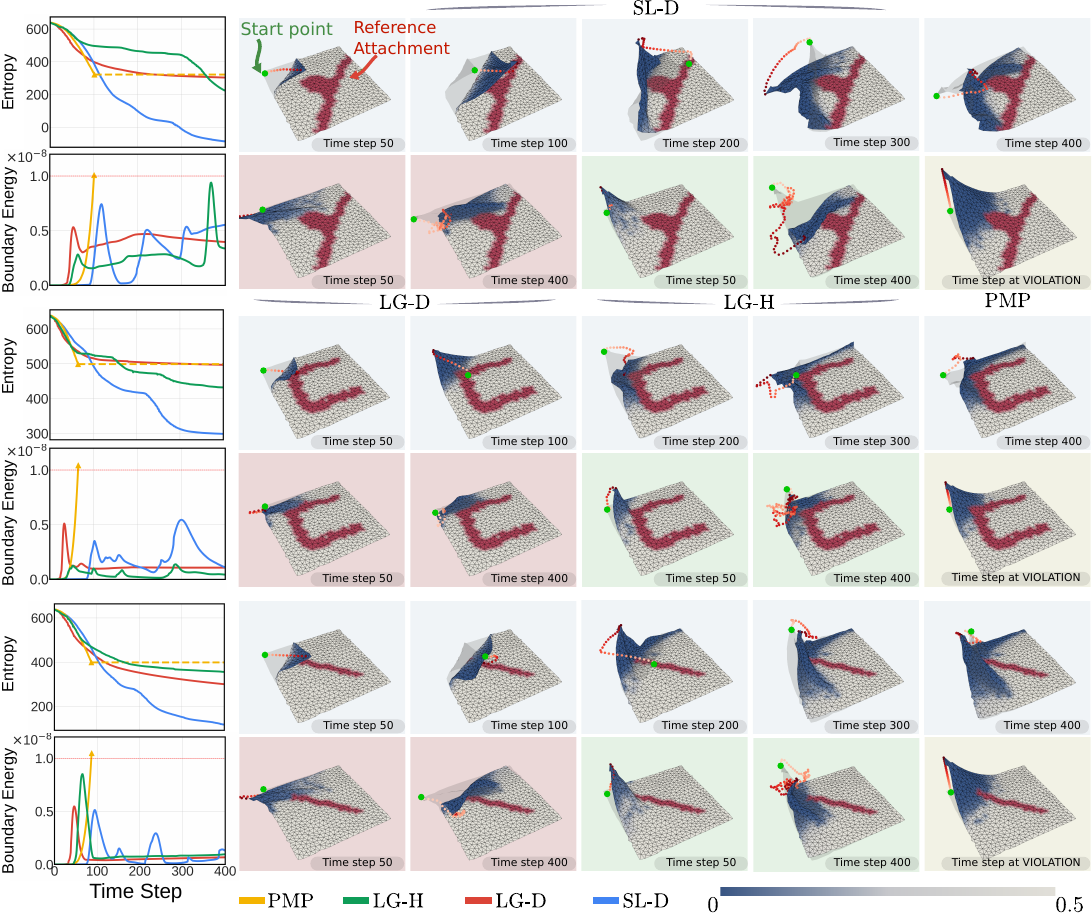}\vspace{-0.7em} 
    \caption{
    Results from our active sensing experiments with 4 different strategies: PMP (yellow) is a baseline, and SL-D (blue), LG-H (green), and LG-D (red) are proposed in this work.
    Every two rows show one experiment with the entropy and energy plotted in the left-most column, and the images on the right show a collage of the control trajectories being applied from the different active sensing strategies.
    Note that the colored background on each image corresponds to the active sensing strategy (best viewed in color).
    The red color on the tissue highlights the reference attachment, and the blue shows the confidence, inverse of variance, of the estimated boundary.
    The goal of the active sensing strategies is to maximize the confidence, which is measured in entropy, while adhering to safety constraints, which are measured in energy.
    Overall, SL-D achieves more entropy reduction than all other baselines while keeping a safe boundary energy profile. 
    It also produces the most intricate control point trajectories, such as switching directions and folding.
    In comparison, local controllers LG-H and LG-D get trapped in local minima, resulting in higher entropy.
    PMP reduces entropy in the beginning but results in quick safety violations.
    }
    \label{fig:active_sensing_graph_trajectory_compare}
    \vspace{-1.5em}
\end{figure*}

\begin{figure*}[t]
    \centering
    \includegraphics[width=1\linewidth]{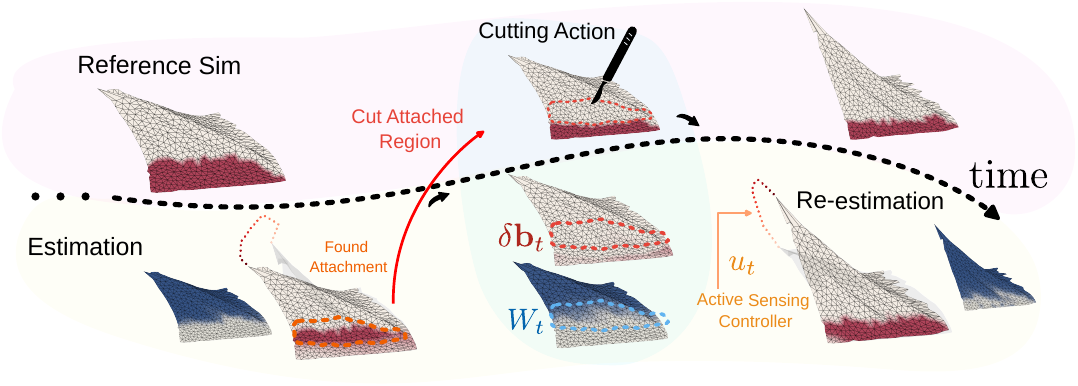} \vspace{-1.5em}
    \caption{
    We applied JIGGLE to a tissue detachment procedure on test case \textit{Large-attach}, which is done by iteratively applying the active sensing approach to find the attachment points, and then a cut is made at that discovered boundary.
    The sequence of images depicts an iteration of this procedure where the top row shows the ground-truth boundary attachment in red, and the bottom row shows our estimation of the boundary in red and the confidence, inverse of variance, in blue.
    After each cut action is applied, where all of the discovered boundary is removed, the cutting information, $\delta \bel_{t}, W_{t}$, is also fed into our estimation algorithm.
    The strategy is repeated until the tissue is fully detached and it took 7 cycles in this experiment to detach the tissue successfully .
    }
    \label{fig:reeestimate_cut}\vspace{-1em}

\end{figure*}



\subsection{Effectiveness of Active Exploration} \label{sec:active_exploration_experiment}

We compare the active sensing performance of four different variants of the proposed framework: 
\begin{enumerate}[leftmargin=0.5cm]
    \item \textbf{LG-H}: Local gradient-based action, minimizing entropy \textcolor{rebuttal}{using Eq.~\ref{eqn:local_entropy_loss} and Eq.~\ref{eqn:local_action_update_entropy},}
    \item \textbf{LG-D}: Local gradient-based, maximizing displacement \textcolor{rebuttal}{using Eq.~\ref{eqn:local_loss_displacement} and Eq.~\ref{eqn:local_action_update_entropy},}
    \item \textbf{SL-D}: Sampling-based large step action, maximizing displacement \textcolor{rebuttal}{using Eq.~\ref{eqn:local_loss_displacement}, Eq.~\ref{eqn:large_step_minimization}, and Eq.~\ref{eqn:large_step_action}, and}
    \item \textbf{PMP}: Predefined motion primitives, a method that conducts an exhaustive search to minimize entropy over motion primitives of $(\pm x, \pm y, \pm z )$. \cite{boonvisut2014identification}
\end{enumerate}
To give the baseline, PMP, the best chance for comparison, we run all motion primitives with our estimation framework and show the result of the motion primitive that gives the most entropy reduction.
All active exploration experiments in this sub-section generate a single continuous control sequence without changing to a new grasp point. 
This mimics surgical scenarios where re-grasping different regions of the tissue may be limited due to constraints in the robot's workspace. 


We desire our framework to output actions that maximize information gain while remaining safe.
We measure information gain as entropy reduction of our estimated boundary parameter belief $\entropy(\bel_{t})$. 
An action is considered safe if the resulting mean boundary potential energy of the system remains below a specified threshold, which we set as $\energy_{b, \max}=1e-8$.
This threshold is our safety constraint, and if violated, represents that damage was caused to the tissue (e.g. tearing).

The simulated environments and results are shown in Fig.~\ref{fig:active_sensing_graph_trajectory_compare}.
The PMP method successfully reduces the entropy of $\bel_{t}$. However, in each experiment, it quickly causes the energy to surge and exceed our safety threshold. 
Although the authors in \cite{boonvisut2014identification} are able to tune their motion primitives to operate safely for their demonstration, in reality, hand-tailoring unique, safe motion primitives for individual environments is infeasible. 
The other control rules, LG-H, LG-D, SL-D, can reduce entropy while staying within the safety boundary. 
While LG-D and LG-H perform comparably, we notice that LG-D is, on average, capable of slightly greater entropy reduction than LG-H. 
We hypothesize that this is due to LG-H's heavy reliance on the estimated belief, which leads it to get stuck in more local minima than LG-D, whose heuristic shows greater detachment from the current boundary parameter estimate. 


When comparing entropy reduction, we see that the local gradient methods LG-H and LG-D were able to achieve entropy reduction comparable to PMP while maintaining a much lower mean boundary energy.
SL-D largely outperformed other methods with improved entropy reduction outcomes which we believe is due to SL-D's ability to avoid local minima.
All our proposed methods, including the local gradient approaches, are capable of complex control strategies. 
Fig. \ref{fig:active_sensing_graph_trajectory_compare} even shows how the robot selects actions that flip and fold uncertain regions in the cloth to get observations that reduce entropy without excessively pulling at the attached regions.

Despite their more rigorous formulation, the local gradient methods failed to decrease entropy much beyond the baseline PMP method.
In Fig.~\ref{fig:active_sensing_graph_trajectory_compare}, both the entropy plots and the variance graph show that these local methods fail to make much progress after the first 100 time steps. 
We hypothesize that this is because these methods are susceptible to getting trapped in the local minima of their complex objective functions.
This is evidenced by viewing the trajectories in Fig. \ref{fig:active_sensing_graph_trajectory_compare}, where in trajectories requiring complex pulling and folding on opposite sides LG-D and LG-H get stuck after reducing entropy by pulling on one side. 
SL-D, however, continues to reduce entropy even after the local methods get stuck by overcoming the gradient locality problem via large-step sampling. 
By sampling and evaluating several long-range trajectories, this method is able to manipulate the cloth in directions that lead to larger entropy reduction in the long term that may not be apparent through short-term local evaluation.

\begin{figure}[t]
    \centering
    \includegraphics[width=0.98\linewidth]{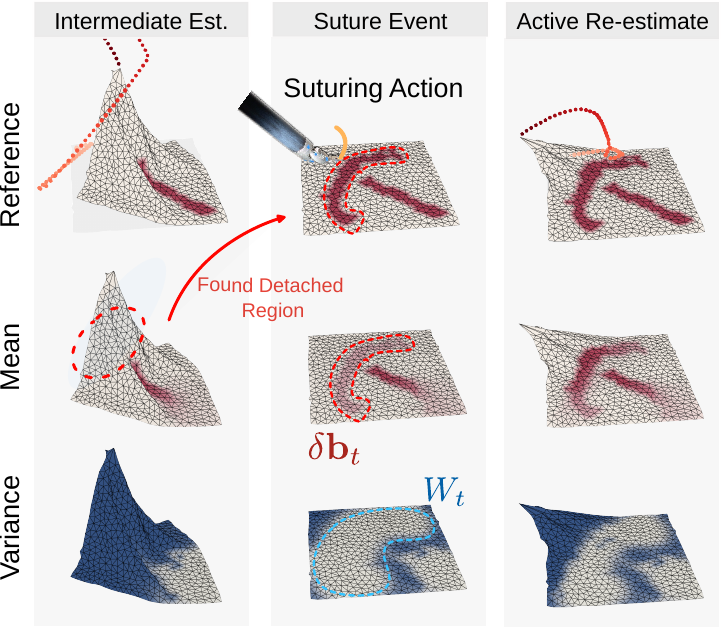} 
    \caption{
    We applied JIGGLE to a suturing procedure by 1) the left column: applying the active sensing approach to find where the tissue is detached,
    2) the middle column: applying a suture action at the desired detached area, and 3) the right column: active sensing again to validate the suture action.
    The top row of images shows the ground truth attachment in red, and the bottom row shows our estimation of the boundary in red and the confidence, inverse of variance, in blue.
    After each suture is applied, the suture information is also fed into our estimation algorithm, including a noise injection near the suture region, so the active sensing policy is encouraged to confirm where the new boundary has been added.
    The second re-estimation, confirming the boundary after the suture, reports an increase in PUG from $20.2$ to $75.2$. 
    The PCD value remains the same at $100$, indicating no false detections.
    }
    \label{fig:reestimate_suture}\vspace{-1.5em}
\end{figure}

\subsection{Adaptation to Topological Changes}\label{sec:reestimation_experiment}
We show how our framework can be used with topology-changing actions like cutting and suturing in tasks that require re-estimation of the boundary parameters. 
We showcase the cutting case in Fig.~\ref{fig:reeestimate_cut}, where the goal is to free the tissue from the environment by iteratively finding and severing all attachment points. 
In this experiment the tissue itself is not cut, but rather the connection between the tissue and the base environment.
The iterative approach first uses our active sensing algorithm to find attachment points on the tissue. 
A cutting action is then used to detach the found region.
Knowledge of the cut is used to update our mean and variance estimations using $\delta \bel_{t}, W_{t}$ and the model $\motionmodel$. 
$\delta \bel_{t}$ is set to be negative near the points selected to be cut with the cutting action. 
$W_{t}$ is set to inject noise into area close to the cut regions, indicating uncertainty about the success of our action. 
Being attentive to the increased uncertainty, our active sensing strategy takes actions to safely minimize entropy and re-estimates the boundary parameters. 
This both validates the effect of our cut action, as well as localizes the remaining attachment points to be cut with the next action.

\begin{figure}[t]
    \centering
    \includegraphics[width=1\linewidth]{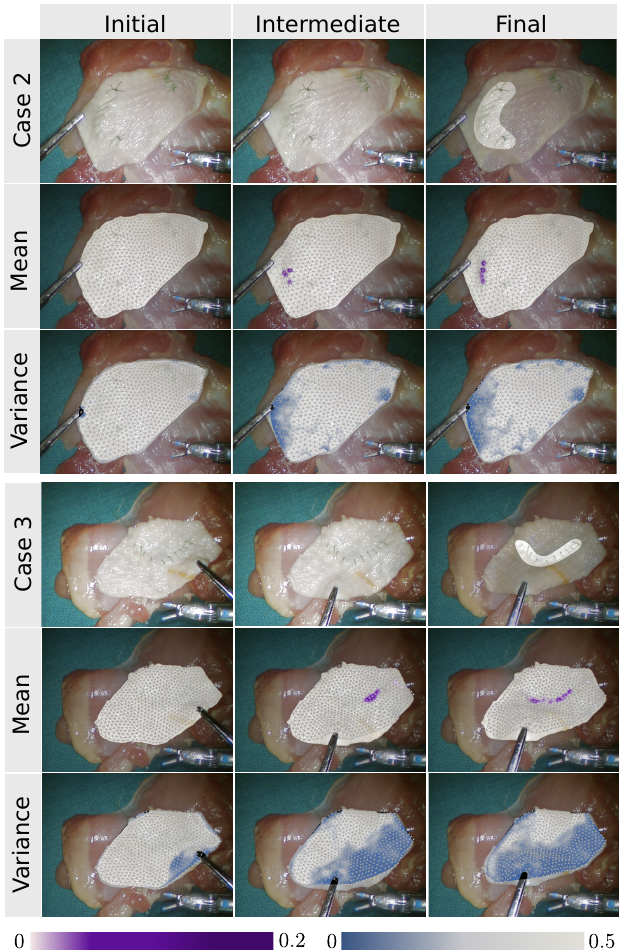} \vspace{-1.5em}
    \caption{
    This figure shows results from real-world tissue attachment point estimation experiments case 2 and case 3 (case 1 is shown in Fig.~\ref{fig:cover_figure}).
    Time progresses from left to right. A dark purple value corresponds to a stronger estimated attachment point, and a darker blue value corresponds to lower uncertainty about the boundary parameters at that region. 
    Noticed that as the tissue is deformed, the variance decreases, and the estimated mean of our boundary matches closer to the ground truth. 
    }
    \label{fig:real_estimation_time}
    \vspace{-1.7em}
\end{figure}
We also show our framework used for suturing tasks in Fig.~\ref{fig:reestimate_suture}. 
We first identify a free region to be attached with a suture (i.e. adding a boundary). 
After a suturing action, we update our boundary parameter estimate in a similar manner as the cutting case. 
$\delta \bel_{t}$ is chosen to be greater than $0$ at values near the sutured region to indicate the added attachment. 
$W_{t}$ is set to inject noise near the sutured region, indicating uncertainty in the success of creating new attachments. 
Our active sensing algorithm takes action to estimate the new boundary parameters, minimizing uncertainty around the modified region and verifying the success of our actions.

\begin{figure}[t]
    \centering
    \includegraphics[width=1\linewidth]{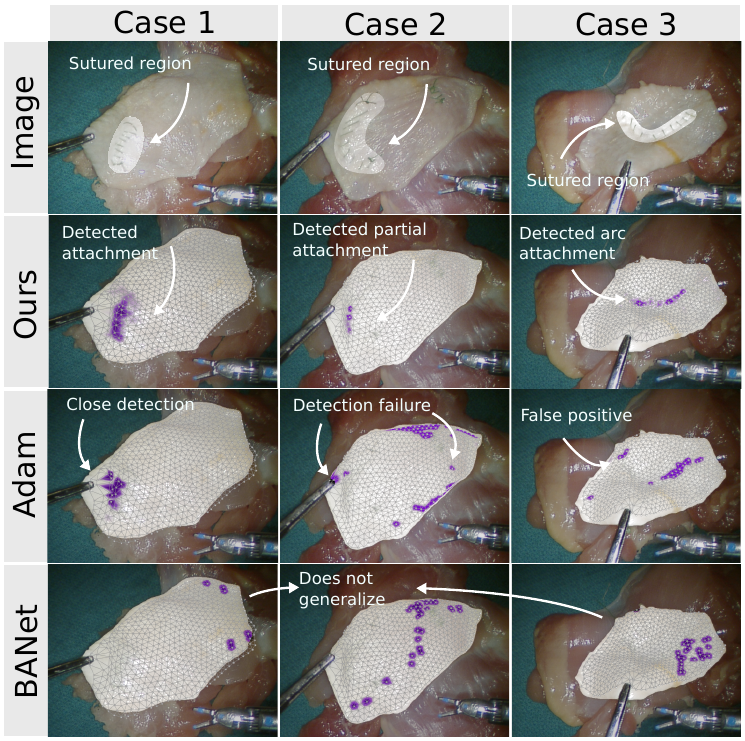} \vspace{-1.5em}
    \caption{Detailed comparison of estimated attachment points between the proposed method and the Adam optimizer. In the second row, our method successfully detects attachment shape. In comparison, the Adam optimizer estimates predictions that are close to the true region (case 1) but fails in other cases, producing many false positive estimations. Because BANet cannot generalize to our real-world data, it fails to predict meaningful results.  
    }\vspace{-0.5em}
    \label{fig:real_estimation_compare}
\end{figure}

\subsection{Estimation Capabilities in Real World}\label{sec:real_world_estimation_accuracy}

Finally, we test our estimation algorithm in a real-world experiment in three cases that mimic surgical tissue manipulation scenarios. 
A layer of chicken skin, which will be manipulated, is placed on top of a larger chicken thigh.
We suture regions of the chicken skin onto the meat to create attachment points.
We use an endoscopic stereo RGB camera from the da Vinci Research Kit (dVRK) \cite{dvrk} for sensing and a Tweezer's tooltip for grasping.
From the collected images, we extract dense surface point cloud observations of the chicken skin tissue deformation. 
We use Segment-Anything \cite{kirillov2023segany} for semantic segmentation and Raft-stereo \cite{lipson2021raft} for disparity estimation. 
We then transform 2D image pixels to 3D using an inverse camera projection model. 
The tweezer's tooltip trajectories are first manually labeled on the images and then transformed to 3D space in a similar manner. 
Initial simulation meshes are reconstructed from the first point cloud observations of the chicken skin.
To reduce the real-to-sim gap that is not caused by boundary connections, we perform system identification on other non-boundary simulation parameters using tools from  \cite{liang2023realtosim}.
In total, 3 test cases are collected on 3 pieces of chicken skins, each with unique attachment shapes. In case 1 and case 2, there is only a single grasp sequence. In case 3, we apply two grasp sequences to the same piece of chicken skin tissue by resetting and re-grasping when the algorithm cannot recover more information from the first grasp point without violating safety constraints.
We again compare against the Adam optimizer baseline, along with \textcolor{rebuttal}{another deep neural network approach, BANet \cite{9359348}.} \textcolor{rebuttal}{The neural network takes in a 3D volume representing the tissues and its deformation field at the last time point of each test cases as inputs, classifying each voxel as boundary or not. We use the pre-trained weights from the original BANet work. More details on our adaptation of BANet is discussed in Appendix \ref{appendix:banet_details}.}
 

\begin{table}[t!]
\setlength\tabcolsep{0.5em}
\centering
\caption{Results from our real world estimation experiment. The number after the metrics corresponds to how much dilation is applied when computing the metric.
}
\begin{adjustbox}{width=0.48\textwidth}
    \begin{tabular}{lcccccc}
    \toprule
  & PCD \upgreenarrow & PCD-1 & PCD-2 &PUG  \upgreenarrow & PUG-1 & PUG-2  \\
  \midrule
Case1-BANet & 0 & 0 & 0 & 0 & 0 & 0 \\
Case1-Adam & 46.2 & $\mathbf{84.6}$& 84.6 & 54.5 & 100.0 &100.0 \\
Case1-Ours & $\mathbf{47.4}$& 84.2& $\mathbf{89.5}$& $\mathbf{81.8}$& 100.0 & 100.0\vspace{0.5 mm}\\\hdashline
Case1-Ours-$\sigma$=0.1 & 60& 93.3& 100& 81.8& 100.0 & 100.0\\
\midrule
Case2-BANet & 0 & 0 & 7.1 & 0 & 0 & 7.5 \\
Case2-Adam& 0.0 & 2.1 & 10.4 &0.0 & 2.5 & 17.5\\
Case2-Ours & $\mathbf{33.3}$& $\mathbf{83.3}$& $\mathbf{100.0}$& $\mathbf{5.0}$ & $\mathbf{27.5}$ & $\mathbf{37.5}$ \vspace{0.5 mm} \\
\hdashline
Case2-Ours-$\sigma$=0.1 & 40.0 & 100& 100.0 & 11.1& 55.6&  72.2\\
\midrule
Case3-BANet & 8.0 & 20.0 & 32.0 & 4.2 & 14.5 & 22.5 \\
Case3-Adam& $\mathbf{45.2}$ & 61.3 & 67.7 & $\mathbf{42.4}$ & $\mathbf{81.8}$ & 97.0\\

Case3-Ours & 44.5 & $\mathbf{72.2}$ & $\mathbf{88.9}$ & 24.2 & 78.8 & $\mathbf{100.0}$ \vspace{0.5 mm}\\
\hdashline
Case3-Ours-$\sigma$=0.1 & 40.0 & 70.0& 100.0 & 14.8&37.0& 63.0\\
    \bottomrule
\end{tabular}
\end{adjustbox}
\label{tbl:quantitative_real_world}
\end{table}

Quantitative evaluations are done using the PCD and PUG metrics with up to two dilation iterations (about 5 mm expansion).
We show the performance of the final estimation of each case in Table \ref{tbl:quantitative_real_world}.
Because our method produces meaningful uncertainty, we also report our method's accuracy of results within a certainty threshold.
We do this by excluding uncertain areas where the variance is larger than 0.1 when computing the accuracy metrics. 
In Table \ref{tbl:quantitative_real_world}, we denote it as ``Ours-$\sigma\text{=}0.1$". The PCD metric becomes better since our method is uncertain about some false positive results, and we are able to exclude them. 
A side effect is that the PUG metric for case 3 becomes lower as some true positive results are uncertain.

We also show example results from our proposed approach in case 1 in Fig. \ref{fig:cover_figure} and case 2 \& 3 in Fig. \ref{fig:real_estimation_time}
In case 2, the attached region is small and close to the grasp point. 
When the tissue is lifted up, the estimated results (mean shown in purple, variance shown in blue) gradually uncover the part of the attached region. 
The final variance is relatively low around the grasping point compared to other not-displaced regions, and the estimated boundary points line up with the true sutures.
In case 3, the attachment region has a larger arc shape and is further from the grasping point. 
Over time, our estimation improves as uncertainty gradually decreases across the tissue, and the mean value of the boundary parameters is correctly identified near the attachment points. 

Our methods outperforms Adam optimizer in terms of PCD, especially in the hardest case 2; this is because our method is less prone to false positives in the estimation, demonstrating the EKF's ability to handle noisy observations. This is illustrated in Fig.~\ref{fig:real_estimation_compare} as more false positive results appear for the Adam optimizer.
For PUG, our method perform comparably to the Adam optimizer but is better in case 2. Shown in Fig. ~\ref{fig:real_estimation_compare}, the comparison method is not able to predict attachment close to the sutured region whereas ours can. BANet performs relatively poorly in our experiments as shown in both Table~\ref{tbl:quantitative_real_world} and the last row of Fig.~\ref{fig:real_estimation_compare}. This is possibly because of its inability to generalize to out-of-distribution targets and the real-to-sim gap between our data and the simulation data that it was trained on.



\subsection{Computation Performance}\label{sec:computation_performance}
We report the average runtime of each component of our method.
In all experiments, tissues are simulated with 600 particles.
The update step of our estimation framework requires $0.25s$ to update belief distribution.
Using the multiple shooting metric with $3$ samples increases the runtime to $0.36s$. 
A single optimization iteration of our local method takes $0.11s$, for LG-D, and $0.8s$, for LG-H. 
SL-D returns an action after $2.9s$ when run for $5$ optimization iterations with $100$ samples. 
These numbers support the use of JIGGLE online. Additionally, parallel and asynchronous computation can be incorporated to achieve further performance gain.

%% file: main_tex_files/discussion_conclusion.tex
In this paper, we proposed \algname: a novel estimation and active sensing framework to identify attachment points in a deformable surgical environment. 
We provide a complete framework for boundary parameter identification from vision to estimation to active control. 
With our probabilistic approach, we were able to provide an estimation method that is robust to the noisiness of real data and an active sensing method to maximize information gain safely.
By providing this robust fully integrated framework and demonstrating its effectiveness even with topological changes, we take large strides towards improving the field of surgical automation. 

Despite our choice of XPBD, our approach is not particular to the choice of differentiable simulator. 
Though we do not explicitly consider test cases where the true strength of the attachment regions varies, our framework supports the estimation of continuous parameters capable of identifying boundary parameters of varying stiffness. 
\textcolor{rebuttal}{While we demonstrate our method on homogenous tissues, with uniform attributes like tissue stiffness, our method is able to handle small amounts of variability in the tissue parameters through the noise term in the EKF. 
This is showcased in our real world experiments as real tissues, like chicken skin, show natural variation in their tissue attributes. 
In future work we seek to further explore the challenges presented by heterogeneous tissues through simultaneous estimation of tissue and boundary parameters. 
}
Finally, while we primarily focus on the surgical use case, our method can be applied to other general thin shell deformables, such as household tasks that involve manipulating cloth. 
In doing so, our method lays the groundwork for a generalizable approach for manipulators to achieve complex cutting and forming of other soft bodies. 

\textcolor{rebuttal}{Though the framework is evaluated on thin-shell subjects, we believe it can potentially work for volumetric deformable objects as well. The major challenge will be decoupling tissues' internal deformation and deformation caused by boundary attachments. To do this, we plan on adding additional active sensing objectives in the future to encourage direct observation of possible attachment regions.} 
Despite the efficacy of our results, the imperfect nature of our estimation stems from the real-to-sim gap \cite{liang2023realtosim} \textcolor{rebuttal}{and discretization errors}. 
\textcolor{rebuttal}{In future works, we also seek to address these to further improve our results through better system identification and model refinement}. 

%% file: appendix.tex



\newpage
\section{Appendix}
\subsection{\textbf{Relationship Between Uncertainty-Weighted Displacement (UWD) and Boundary Entropy}} \label{appendix:relationship_displacement_entropy}
    In this section we present the following proposition, and provide a proof tying the relationship between the UWD, i.e., displacement, $\weighteddeform$, of the tissue particles from their initial position, and the reduction in boundary entropy $\entropy(\mathbf{b_{t+1}})$ at the next timestep $t+1$. 

        \begin{proposition}
            The active sensing objective of maximizing the uncertainty-weighted deformation displacement, i.e.,
                $\weighteddeform$ in Eq. \ref{eqn:uncertain_weighted_displacement},
            is designed to decrease the entropy $\entropy(\mathbf{b}_{t+1})$ of boundary estimation and uncover more unknown information. That is,
                \begin{equation} \label{eqn:displacement_vs_entropy}
                    \begin{split}
                        \min \entropy(\mathbf{b}_{t+1}) \propto \max \weighteddeform
                    \end{split}
                \end{equation}

        \end{proposition}
        
    To prove it, we first establish a connection to entropy, $\entropy$, via the observation Jacobian $\beljacob_{t+1} = \frac{\partial \obsmodel}{\partial \bel}$, defined in Eq. \ref{eqn:kalman_jacobians} of the EKF equations in section \ref{appendix:entropy_to_jacobian}. 
    Second, we derive the relationship between the Jacobian $\beljacob$ and the tissue displacement in section \ref{appendix:jacobian_to_uwd}. 
    Third, we combine these results to show proposition 1 in section \ref{appendix:entropy_to_uwd}. 

\subsubsection{\textbf{Boundary Entropy Regarding the Observation Jacobian}}\label{appendix:entropy_to_jacobian}
    As defined in Eq. \ref{eqn:kalman_entropy}, it is evident that the entropy $\entropy(\mathbf{b}_{t+1})$ is directly proportional to the determinant of the covariance matrix outputted by the Extended Kalman Filter (EKF) estimator. 
    Since $\Sigmab_{t|t}, \motioncov_{t}, \obscov_{t}$ are all positive semi-definite covariance matrices, they allow the following factorization: 
    \begin{equation}\label{eqn:roots_Sigmab}
        \begin{split}
            \Sigmab_{t+1|t} & = \Sigmab_{t|t} + \motioncov_{t} = \Sigmab^{\frac{1}{2}} \Sigmab^{\frac{1}{2}}
        \end{split}
    \end{equation}
    Combining this fact with the Extended Kalman Filter (EKF) equations, as defined in Eq. \ref{eqn:predict_step} and \ref{eqn:update_step}, and for the sake of simplicity, neglecting the subscripts of observation Jacobian $\beljacob_{t+1}$ and Kalman gain $\kalmangain_{t+1}$, we obtain:
    \begin{equation}
        \label{eqn:entropy_beljacob}
        \begin{split}
            \entropy(\mathbf{b}_{t+1}) 
            & \propto \Big| \Sigmab_{t+1|t+1} \Big| \\
            & \propto \Big| \Big( I - \kalmangain \beljacob \Big) \Sigmab_{t+1|t} \Big| \\
            & \propto \Big| \Big( I - \Sigmab_{t+1|t} \beljacob^{\top}
                \big[\beljacob \Sigmab_{t+1|t} \beljacob^{\top} + \obscov_{t}\big]^{-1} \beljacob \Big) \Sigmab_{t+1|t}  \Big| \\
            & \propto \Big| \Big( I - \Sigmab^{\frac{1}{2}} \Sigmab^{\frac{1}{2}} \beljacob^{\top}
                \big[\beljacob \Sigmab^{\frac{1}{2}} \Sigmab^{\frac{1}{2}} \beljacob^{\top} + \obscov_{t}\big]^{-1} \beljacob \Big) \Sigmab^{\frac{1}{2}} \Sigmab^{\frac{1}{2}} \Big| \\
            & \propto \Big| \Big( I - \Sigmab^{\frac{1}{2}} \beljacob^{\top}
                \big[\beljacob \Sigmab^{\frac{1}{2}} \Sigmab^{\frac{1}{2}} \beljacob^{\top} + \obscov_{t}\big]^{-1} \beljacob \Sigmab^{\frac{1}{2}} \Big) \Big| \Big|\Sigmab^{\frac{1}{2}} \Sigmab^{\frac{1}{2}} \Big| \\
        \end{split}
    \end{equation}


    Denoting $\mathsf{A} = \beljacob \Sigmab^{\frac{1}{2}}$ and $\Sigmab_{t+1|t} = \Sigmab = \Sigmab^{\frac{1}{2}} \Sigmab^{\frac{1}{2}}$, we can perform singular value decomposition (SVD)  
    where $D$ is a matrix with singular values $\singularval$ on its diagonal, $UU^{T} = U^{T}U = I$, and $QQ^{T} = Q^{T}Q = I$. 
    \begin{equation}
        \label{eqn:SVD}
        \begin{split}
            \mathsf{A}  & = \beljacob \Sigmab^{\frac{1}{2}} = U D Q^{\top}
        \end{split}
    \end{equation}

    For simplification and to maintain a straightforward representation of observation uncertainty, we assume $\obscov_{t} = \alpha I$, with $\alpha$ denotes the maximum expected observation noise variance. 
    This diagonal covariance matrix implies that the observation noise is isotropic and of uniform magnitude across different dimension. Substituting them into Eq. \ref{eqn:entropy_beljacob}, we get: 
    \begin{equation} \label{eqn:entropy_beljacob_singularvals}
        \begin{split}
            \entropy(\mathbf{b}_{t+1}) & \propto \\
            & \propto \Big|I -  \mathsf{A}^{\top}\big[\mathsf{A}\mathsf{A}^{\top} + \alpha I\big]^{-1} \mathsf{A} \Big| \Big|\Sigmab\Big| \\
            & \propto \Big|QQ^{\top} - QD^{\top}U^{\top} \big[UDD^{\top}U^{\top} + \alpha I\big]^{-1} UDQ^{\top} \Big| \Big|\Sigmab\Big|\\
            & \propto \Big|Q\Big| \Big|I - D^{\top}U^{\top} \big[UDD^{\top}U^{\top} + \alpha I\big]^{-1} UD \Big| \Big|Q^{\top}\Big| \Big|\Sigmab\Big| \\
            & \propto \Big|Q\Big| \Big|Q^{\top}\Big| \Big|I - D^{\top}U^{\top} \big[UDD^{\top}U^{\top} + \alpha I\big]^{-1} UD \Big| \Big|\Sigmab\Big| \\
            & \propto \Big|I - D^{\top}U^{\top} \big[UDD^{\top}U^{\top} + \alpha I\big]^{-1} UD \Big| \Big|\Sigmab\Big| \\
            & \propto \Big|I - D^{\top}U^{\top} \big[UDD^{\top}U^{\top} + \alpha UU^{\top}\big]^{-1}UD \Big| \Big|\Sigmab\Big| \\
            & \propto \Big|I - D^{\top}U^{\top} U \big[DD^{\top} + \alpha I\big]^{-1} U^{\top}UD \Big| \Big|\Sigmab\Big|\\
            & \propto \Big|(I - D^{\top}(DD^{\top} + \alpha I)^{-1} D)\Big| \Big|\Sigmab\Big|\\
            & \propto |\Sigmab| \prod_{i}^{N} \Big(1 - \frac{\singularval^{2}}{\singularval^{2} + \alpha}\Big) 
        \end{split}
    \end{equation}
    In Eq. \ref{eqn:entropy_beljacob_singularvals} and Eq. \ref{eqn:roots_Sigmab}, the only term influenced by our control actions at time $t$ is the observation Jacobian $\beljacob$, as $\Sigmab$ only depends on the estimation from the previous step. These equations show that selecting $\beljacob$ in a manner that increases the magnitudes of the singular values $||\singularval||_{2}$ of $\beljacob \Sigmab^{\frac{1}{2}}$ will result in a reduction of the entropy within the belief distribution.

    After establishing the correlation between the Jacobian and entropy reduction, we need to link the uncertainty-weighted displacement (UWD) of particles to the Jacobian. We must identify the choices of UWD that lead to entropy reduction. Demonstrating these connections will validate that our Proposition \ref{eqn:displacement_vs_entropy}.

    \begin{figure}[!tb]
        \centering
        \includegraphics[width=0.6\linewidth]{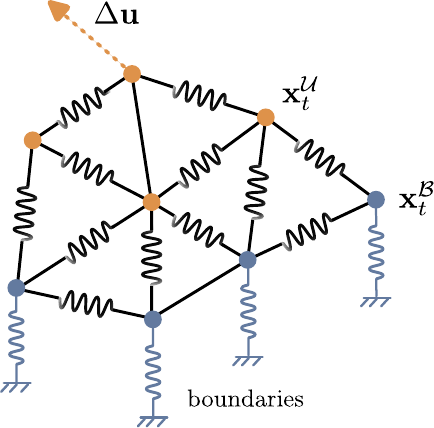} 
        \caption{\textcolor{black}{
        This figure shows a simplified model of the XPBD simulation. 
        The particles shown in blue, $\boundarypt$, are related to the boundary constraints. 
        We are trying to estimate the boundary parameters associated with the boundary constraints on these particles. 
        The remaining particles are shown in orange, $\otherpt$.
        The virtual particle $\action$ is used to apply control on the tissue. 
        All constraints are implemented as set distance constraints and are depicted using springs with different springs constants set according to the associated constraint. 
        }}
        \label{fig:pbd_structure_proof}
    \end{figure}
    
    \subsubsection{\textbf{Observation Jacobian Regarding the Uncertainty-Weighted Displacement (UWD)}}\label{appendix:jacobian_to_uwd}
    To link the Jacobian, $\beljacob$, at the next timestep $t+1$, to displacement we derive an analytical form of $\beljacob$. 
    To derive the exact form of observation Jacobian, we illustrate the simulated mesh with a simplified graph in Fig. \ref{fig:pbd_structure_proof}. Hence, we partition our whole tissue state representation $\mathbf{x} = \{\boundaryset, \otherset\}$ into particles related to boundary constraints $\boundaryset$ and other particles denoted as $\otherset$. 
    Of the $n$ total particles, we assume that there are $m$ particles in $\boundaryset$ and $q$ particles in $\otherset$ such that $m + q = n$. 
    From the definition in Eq. \ref{eqn:elastic_potential}, the XPBD simulation involves minimizing the total energy potential of constraints:
    \begin{equation} \label{eqn:energy_minimization_xpbd_into_parts}
    \begin{split}
        \arg\min_{\mathbf{x}} \energy(\mathbf{x}, \mathbf{b}) = \arg\min_{\mathbf{x}} \energy_1(\boundaryset, \mathbf{b}) + \energy_2(\otherset, \boundaryset)
    \end{split}
    \end{equation}
   where the first term, $\energy_1(\otherset, \boundaryset)$, representing the energy only related to the current boundary constraints, $\mathbf{b}$ and $\energy_2(\otherset, \boundaryset)$ denotes the energy term associated with both $\otherset$ and $\boundaryset$.
   In our XPBD simulator, all defined constraints are distance constraints between particles.
   For the boundary energy, we are able to write: 
    \begin{equation} \label{eqn:energy_U1}
    \begin{split}
        \energy_1(\boundaryset, \mathbf{b})  
        & = \frac{1}{2} 
        \mathbf{C}^{\top}(\boundaryset)\mathtt{diag}(\mathbf{b})\mathbf{C}(\boundaryset) \\
        & = \frac{1}{2} \sum_{i=1}^m b_i \Big(\norm{\mathbf{x}^{\mathsf{B}, i}_{t+1} - \mathbf{x}^{\mathsf{B}, i}_{0}}_{2} - d_{0}^{\mathsf{B}, i} \Big)^2 
    \end{split}
    \end{equation} 
    $\mathbf{b} \in \mathbb{R}^{m}$ where $b_i$ is the stiffness of the $i$-th boundary spring, of the total $m$ springs.
    $\boundarypt^{,i} - \mathbf{x}^{\mathsf{B}, i}_{0}$ is the displacement of particle $i$ related to the boundary constraint from time $0$ to the next time step $t+1$. 
    $d_{0}^{\mathsf{B}, i}$ is the length of the boundary spring associated with particle $i$ at rest time, $t=0$. 
    For our work, we consider a zero length spring at rest, $d_{0}^{\mathsf{B}, i} = 0, \forall i$, for all boundary springs. 
    Similarly, we describe the rest of spring potential energy by: 
    \begin{equation} \label{eqn:energy_U2}
    \begin{split}
        \energy_2(\otherset, \boundaryset)  
        & = \frac{1}{2} 
        \mathbf{C}^{\top}(\otherset, \boundaryset)\mathtt{diag}(\mathbf{k})\mathbf{C}(\otherset, \boundaryset) \\
        & = \frac{1}{2} \sum_{j=1}^s k_j \Big(\norm{\mathbf{x}^{j,1}_{t+1} - \mathbf{x}^{j,2}_{t+1}} - d^{1,2,j}_{0} \Big)^2 
    \end{split}
    \end{equation}
    where $\mathbf{x}^{j,1}_{t+1}, \mathbf{x}^{j,2}_{t+1} \in \{\boundaryset \cup \otherset\}$ are a pair of two particles connected by the $j$-th spring within the simulated mesh.  
    $d^{1,2,j}_{0}$ is the distance of that $j$-th spring constraint between particle $1$ and $2$ at rest, at time $t=0$. 
    Using this, we are able to define the matrices $\mathbf{L} \in \mathbb{R}^{3n\times3n}$ and $\mathbf{P} \in \mathbb{R}^{3n\times3s}$ as follows:
     \begin{equation} \label{eqn:matrices_L_J}
    \begin{split}
         \mathbf{L} & = \bigg(\sum_{j=1}^s k_j \mathbf{A}^j \mathbf{A}^{j, \top} \bigg) \otimes \mathbf{I}_3 \\
         \mathbf{P} & = \bigg(\sum_{j=1}^s k_j \mathbf{A}^j \mathbf{S}^{j, \top} \bigg) \otimes \mathbf{I}_3 \\
         \mathbf{d} & = \frac{\mathbf{x}^{j,1}_{0} - \mathbf{x}^{j,2}_{0}}{ \norm{\mathbf{x}^{j,1}_{0} - \mathbf{x}^{j,2}_{0}} }_{2} d^{1,2,j}_{0} \\
    \end{split}
    \end{equation}
    Here, $\mathbf{A}_j \in \mathbb{R}^{n}$ is the connectivity vector of the $j$-th spring, i.e., $\mathbf{A}^{j,1}=1$, $\mathbf{A}^{j,2}=-1$, and zero otherwise. 
    $(j,1)$ and $(j, 2)$ denote the indices of the first and second particle, respectively, associated with the $j$-th spring constraint. 
    $\mathbf{S}^j \in \mathbb{R}^{s}$ is $j$-th spring indicator, i.e.,  $\mathbf{S}^{j}=\mathbf{1}_s$, $s$ is the number of non-boundary related springs or edges, and $\otimes$ denotes Kronecker product. 
    $\mathbf{d}$ can be interpreted as a spring in the rest pose, at time $t=0$, that is rotated in a specific direction, effectively behaving like a constant. 
    Eq. \ref{eqn:energy_U1} can be re-factorized into: 
    \begin{equation} \label{eqn:energy_U1}
    \begin{split}
        \energy_2(\otherset, \boundaryset)  
        & = \frac{1}{2} \mathbf{x}^{\top} \mathbf{L} \mathbf{x} - \mathbf{x}^{\top} \mathbf{P} \mathbf{d}
    \end{split}
    \end{equation}
    The solution to the minimization function described in Eq. \ref{eqn:energy_minimization_xpbd_into_parts} can be regarded to satisfy:
    \begin{equation} \label{eqn:energy_U1_U2_soluation}
    \begin{split}
        \nabla_{\mathbf{x}} \energy(\mathbf{x}, \mathbf{b})= \nabla_{\mathbf{x}} \energy_1(\boundaryset, \mathbf{b}) + \nabla_{\mathbf{x}} \energy_2(\otherset, \boundaryset) = 0
    \end{split}
    \end{equation}

    Before moving forward we explicitly clarify the structure of our state $x$. 
    Assuming a system with $2$ particles, $\boundaryset = [\boundaryset[0], \boundaryset[1], \boundaryset[2]]^\top$ is the single particle related to boundary constraints and $\otherset = [\otherset[0], \otherset[1], \otherset[2]]^\top$ is the single other particle. 
    $\mathbf{x}$ is a flattened joint state of the $2$ particles in the following form: 
    $\mathbf{x} = [\boundaryset[0], \boundaryset[1], \boundaryset[2], \otherset[0], \otherset[1], \otherset[2]]^{\top}$. 
    For multiple particles we follow a similar structure where the flattened form of the boundary related particles is stacked above the flattened form of the other particles. 
    This gives us: $\boundaryset \in \mathbb{R}^{3m}$, $\otherset \in \mathbb{R}^{3q}$, $\mathbf{x} \in \mathbb{R}^{3n}$

    \textbf{
    \begin{lemma}\textbf{[Implicit-Function Theorem]}
        Considering the solution function defined in Eq. \ref{eqn:energy_U1_U2_soluation},
        \begin{equation} \label{eqn:pbd_implicit_func}
        \begin{split}
             \mathcal{L}(\mathbf{x}^*, \mathbf{b}) = \nabla_{\mathbf{x}} \energy(\mathbf{x}^*, \mathbf{b}) = 0
        \end{split}
        \end{equation}
        with $\mathbf{x}^*$ defined as an equilibrium point, the sensitivities of the solution with respect to the parameter $\mathbf{b}$ can be can be computed as:
        \begin{equation} \label{eqn:pbd_implicit_func}
        \begin{split}
             \frac{\partial \mathbf{x}^*}{\partial \mathbf{b}} = - \bigg( \frac{\partial \mathcal{L}}{\partial \mathbf{x}} \bigg)^{-1} \frac{\partial \mathcal{L}}{\partial \mathbf{b}}
        \end{split}
        \end{equation}
    \end{lemma}
    }

    In our EKF formulation, the observation Jacobian is given by
    \begin{equation} \label{eqn:pbd_obs_jacobian}
        \begin{split}
             \beljacob & = \frac{\partial \obsmodel}{\partial \bel} = \frac{\partial \pbd(\mathbf{x}^{\text{ref}}, \action, \bel)}{\partial \bel} = \frac{\partial \mathbf{x}}{\partial \bel}
        \end{split}
        \end{equation}
    Combining it with the implicit-function theorem in Eq. \ref{eqn:pbd_implicit_func}, we obtain the Jacobian as,
    \begin{equation} \label{eqn:jacobian_implicit_func}
        \begin{split}
             \beljacob & = - \bigg( \frac{\partial \mathcal{L}}{\partial \mathbf{x}} \bigg)^{-1} \frac{\partial \mathcal{L}}{\partial \mathbf{b}} = - \bigg( \frac{\partial \nabla_{\mathbf{x}} \energy}{\partial \mathbf{x}} \bigg)^{-1} \frac{\partial \nabla_{\mathbf{x}} \energy}{\partial \mathbf{b}} \\
             & = - \bigg( \frac{\partial^2 \energy}{\partial \mathbf{x} \partial \mathbf{x}^{\top}} \bigg)^{-1} \frac{\partial^2 \energy}{\partial \mathbf{b}\partial \mathbf{x}^{\top}}
        \end{split}
    \end{equation}
    Given $\mathbf{x}= \{\boundaryset, \otherset\}$, we can computer the following gradients as, 
    \begin{equation} \label{eqn:jacobian_implicit_func}
        \begin{split}
            \frac{\partial \energy_1}{\partial \mathbf{x}} &= \begin{bmatrix}
                \Delta \boundaryset^{\top} \mathtt{diag}(\mathbf{b}) & \mathbf{0}\\
            \end{bmatrix} \in \mathbb{R}^{3n \times 1} \\[0.5em]
            \frac{\partial \energy_2}{\partial \mathbf{x}} &= \begin{bmatrix}
                \boundaryset^{\top} \mathbf{L}_{\mathsf{B}\mathsf{B}} + \otherset^{\top} \mathbf{L}_{\mathsf{U}\mathsf{B}} \\
                \otherset^{\top} \mathbf{L}_{\mathsf{U}\mathsf{U}} + \boundaryset^{\top} \mathbf{L}_{\mathsf{B}\mathsf{U}}
            \end{bmatrix}^{\top} \\
            &- (\mathbf{P} \mathbf{d})^{\top} \\[0.5em]
            &\in \mathbb{R}^{3n \times 1}
            \\
            \frac{\partial^2 \energy_1}{\partial \mathbf{x} \partial \mathbf{x}^{\top}} &= \begin{bmatrix}
                \mathtt{diag}(\mathbf{b}) & \mathbf{0} \\
                \mathbf{0} & \mathbf{0} \\
            \end{bmatrix} \in \mathbb{R}^{3n \times 3n} \\[0.5em]
            \frac{\partial^2 \energy_2}{\partial \mathbf{x} \partial \mathbf{x}^{\top}} &= \begin{bmatrix}
                \mathbf{L}_{\mathsf{B}\mathsf{B}}  & \mathbf{L}_{\mathsf{B}\mathsf{U}} \\
                \mathbf{L}_{\mathsf{U}\mathsf{B}}  & \mathbf{L}_{\mathsf{U}\mathsf{U}}
            \end{bmatrix} = \mathbf{L} \in \mathbb{R}^{3n \times 3n} \\[0.5em]
            \frac{\partial^2 \energy_1}{\partial \mathbf{b}\partial \mathbf{x}^{\top}} &= \begin{bmatrix}
                \wideparen{\Delta\boundaryset} \\
                \mathbf{0}
            \end{bmatrix} \in \mathbb{R}^{3n \times m} \\[0.5em]
            \frac{\partial^2 \energy_2}{\partial \mathbf{b}\partial \mathbf{x}^{\top}} &= \mathbf{0} \in \mathbb{R}^{3n \times m}
        \end{split}
    \end{equation}
    \begin{equation}\label{eqn:displacement_in_partials}
        \begin{split}
            \Delta\boundaryset &= \big(\boundaryset - \mathbf{x}^{\mathsf{B}}_{0} \big) \in \mathbb{R}^{3m \times 1} \\
            \wideparen{\Delta\boundaryset} &= \sum_i^m \mathbf{e}_i \mathbf{e}^{\top}_i \otimes [\Delta \boundaryset]_i \in \mathbb{R}^{3m \times m}
        \end{split}
    \end{equation}
    $\Delta\boundaryset$ is the displacement of all boundary related particles from time $0$ to time $t$. 
    $\wideparen{\Delta\boundaryset}$ is a concatenated matrix of the displacement of all boundary related particles. 
    $\mathbf{e}_i$ is the $i$-th column vector of the standard basis of $\mathbb{R}^m$, and $[\Delta \boundaryset]_i \in \mathbb{R}^{3 \times 1}$ be the deformation displacement of the $i$-th boundarry related particle from $\Delta\boundaryset$.
    Note that this is the same expression from the UWD. 
    It should be noticed that $\mathbf{L}$ is symmetric and positive semi-definite.
    Given these partials we get: 
    \begin{equation}\label{eqn:energy_jacobians}
        \begin{split}
            \frac{\partial^2 \energy}{\partial \mathbf{b}\partial \mathbf{x}^{\top}} &= \frac{\partial^2 \energy_1}{\partial \mathbf{b}\partial \mathbf{x}^{\top}} + \frac{\partial^2 \energy_2}{\partial \mathbf{b}\partial \mathbf{x}^{\top}} \\
            \frac{\partial^2 \energy}{\partial \mathbf{x}\partial \mathbf{x}^{\top}} &= \frac{\partial^2 \energy_1}{\partial \mathbf{x}\partial \mathbf{x}^{\top}} + \frac{\partial^2 \energy_2}{\partial \mathbf{x}\partial \mathbf{x}^{\top}} 
        \end{split}
    \end{equation}

    Substituting the above into equation \ref{eqn:pbd_implicit_func}, we get an expression for the observation jacobian: 
    \begin{equation}\label{eqn:jacobian_implicit_final}
        \begin{split}
            \beljacob &= -\begin{bmatrix}
                \mathbf{L}_{\mathsf{B}\mathsf{B}} + \mathtt{diag}(\mathbf{b}) & \mathbf{L}_{\mathsf{B}\mathsf{U}} \\
                \mathbf{L}_{\mathsf{U}\mathsf{B}}  & \mathbf{L}_{\mathsf{U}\mathsf{U}}
            \end{bmatrix}^{-1} 
            \begin{bmatrix}
                \wideparen{\Delta\boundaryset} \\
                \mathbf{0}
            \end{bmatrix} \in \mathbb{R}^{3n \times m}
        \end{split}        
    \end{equation}

    The first term in this expression is a function of the current boundary estimate $\mathbf{b}$, but is constant with respect to the particle states $\mathbf{x}_{t}$. 
    We refer to this term as $\mathcal{R}(\mathbf{b})$:
    \begin{equation}\label{eqn:jacobian_implicit_constant}
        \begin{split}
            \mathcal{R}(\mathbf{b}) &= -\begin{bmatrix}
                \mathbf{L}_{\mathsf{B}\mathsf{B}} + \mathtt{diag}(\mathbf{b}) & \mathbf{L}_{\mathsf{B}\mathsf{U}} \\
                \mathbf{L}_{\mathsf{U}\mathsf{B}}  & \mathbf{L}_{\mathsf{U}\mathsf{U}}
            \end{bmatrix}^{-1} 
        \end{split}        
    \end{equation}
    For the ease of notation we will denote that $\mathbf{D} = \begin{bmatrix}
    \wideparen{\Delta\boundaryset} &
    \mathbf{0}
    \end{bmatrix}^{T}$. 
    Writing $\beljacob = \mathcal{R}(\mathbf{b}) \mathbf{D}$ we can see that the jacobian given a specific boundary estimate $\mathbf{b}$ is a direct function of the displacement. 
    
\subsubsection{\textbf{Boundary entropy Regarding the UWD}}\label{appendix:entropy_to_uwd}
From section \ref{appendix:entropy_to_jacobian} and section \ref{appendix:jacobian_to_uwd}, we see that to minimize entropy we have to maximize the singular values, $\singularval^{2}$, of $\mathcal{R}(\bel_{t}) \mathbf{D} \Sigmab^{\frac{1}{2}}$. 
Where $\mathcal{R}(\bel_{t})$ is a constant given the belief at time $t$, and $\Sigma^{\frac{1}{2}}$ is a constant at time $t$. 

The Frobenius norm of a matrix $\norm{\mathcal{R}(\bel_{t}) \mathbf{D} \Sigmab^{\frac{1}{2}}} = \sqrt{\sum_{i}^{n} \singularval^{2}}$. 
Thus maximizing the frobenius norm is equivalent to maximizing the magnitude of the $\singularval^{2}$ and decreasing entropy. 
Leveraging the submultiplicative property of the frobenius norm, the known invertibility of matrix $\mathcal{R}(\bel_{t})$, and that $\mathcal{R}(\bel_{t}), \Sigma^{\frac{1}{2}}$ are constant matrices we can write: 
\begin{equation}\label{eqn:norm_lower_bound_derivation}
    \begin{split}
        \norm{\mathcal{R}(\bel_{t})^{-1} \mathcal{R}(\bel_{t}) \mathbf{D} \Sigmab^{\frac{1}{2}}\Sigmab^{\frac{1}{2}}} 
        &\leq \norm{\mathcal{R}(\bel_{t})^{-1}} \norm{\mathcal{R}(\bel_{t}) \mathbf{D} \Sigmab^{\frac{1}{2}}\Sigmab^{\frac{1}{2}}} \\
        \norm{\mathbf{D} \Sigmab} 
        &\leq \norm{\mathcal{R}(\bel_{t})^{-1}} \norm{\mathcal{R}(\bel_{t}) \mathbf{D} \Sigmab^{\frac{1}{2}} \Sigmab^{\frac{1}{2}}} \\
        \frac{\norm{\mathbf{D} \Sigmab}}{\norm{\mathcal{R}(\bel_{t})^{-1}}} &\leq \norm{\mathcal{R}(\bel_{t}) \mathbf{D} \Sigmab^{\frac{1}{2}}} \norm{ \Sigmab^{\frac{1}{2}}} \\
        \frac{\norm{\mathbf{D} \Sigmab}}{\norm{\mathcal{R}(\bel_{t})^{-1}}  \norm{ \Sigmab^{\frac{1}{2}}}} &\leq \norm{\mathcal{R}(\bel_{t}) \mathbf{D} \Sigmab^{\frac{1}{2}}} \\
        \eta \norm{\mathbf{D} \Sigmab} &\leq \norm{\mathcal{R}(\bel_{t}) \mathbf{D} \Sigmab^{\frac{1}{2}}} \\
        \eta &= \frac{1}{\norm{\mathcal{R}(\bel_{t})^{-1}}  \norm{ \Sigmab^{\frac{1}{2}}}}
    \end{split}
\end{equation}
Note that the Uncertainty weighted displacement (UWD) is defined as $\weighteddeform = \norm{\mathbf{D} \Sigmab}$.  

Combining Eq. \ref{eqn:entropy_beljacob_singularvals}, Eq. \ref{eqn:jacobian_implicit_final} and  Eq. \ref{eqn:norm_lower_bound_derivation}, we show that at a given timestep $t$, by maximizing this lower bound $\norm{\mathbf{D} \Sigmab}$ we can maximize $\mathcal{R}(\bel_{t}) \mathbf{D}\Sigma^{\frac{1}{2}}$ and thus increase the magnitude of the singular values $\singularval^{2}$ and decrease entropy. 

With this we have shown proposition 1: The active sensing objective of maximizing UWD is designed to decrease entropy $\entropy(\bel_{t+1})$.

\begin{figure}[t]
    \centering
    \includegraphics[width=1\linewidth]{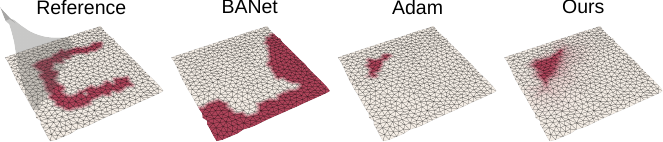}
    \caption{Estimated attachment regions of different method on test case \textit{U-shape}. BANet is trained to estimate hard boundary conditions therefore does not generalize to our data.
    }
    \label{fig:app_banet_compare}
\end{figure}

\subsection{Entropy loss implementation Details}

\begin{figure*}[t]
    \centering
    \includegraphics[width=1\linewidth]{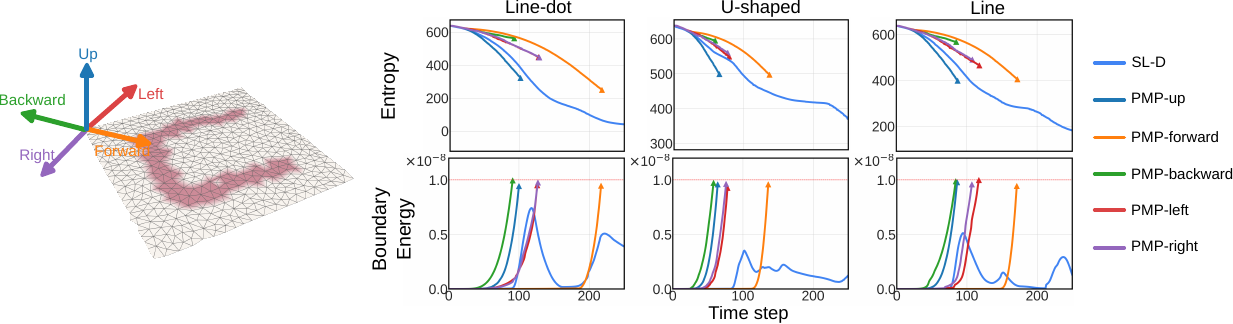}
    \caption{Additional comparison of active sensing performance between all choices of predefined motion primitives (PMP). They are \textit{up}, \textit{forward}, \textit{backward}, \textit{left}, \textit{right}. Results from our proposed Sampling-based Large step controller (SL-D) are included for comparsion. All PMPs result in safety violation. Before violation, \textit{up} and \textit{forward} achieve more entropy reduction, but \textit{up} achieves that faster in general. In comparsion, the proposed method outperform all PMPs, achieving more entropy reduction at the cost of smaller boundary energy.
    }
    \label{fig:app_active_sensing_pmps}
\end{figure*}
In this section we discuss the implementation details for the entropy based loss from Eq.~\ref{eqn:local_entropy_loss}. 
This loss is optimized in our active sensing algorithm, LG-H, to solve for the best next action. 

The entropy of our belief is specified in Eq. \ref{eqn:kalman_entropy} as : 
\begin{equation} 
    \begin{split}
        \entropy(\bel_{t+1}) 
            & \propto \ln\left(|\Sigmab_{t+1}| \right) 
    \end{split}
\end{equation}
The $ln$ makes directly optimizing this quantity in our loss function challenging. 
As uncertainty decreases and $|\Sigmab_{t+1}|$ approaches $0$, $ln(|\Sigmab_{t+1}|)$ tends to $-\inf$. 
This makes it hard to set the weights to balance this loss with the other metrics from Eq.~\ref{eqn:local_entropy_loss} during our unconstrained optimization. 

As a result instead of directly minimizing entropy, $\entropy(\bel_{t+1})$, we minimize the term: 
\begin{equation}\label{eqn:true_entropy_loss}
    \begin{split}
        |I - K_{t+1}\beljacob_{t+1}|
    \end{split}
\end{equation}
From equations Eq. \ref{eqn:kalman_entropy} and Eq. \ref{eqn:update_step} we get that: $\entropy(\mub_{t+1}) \propto \ln\left(|(I - K_{t+1}\beljacob_{t+1})\Sigmab_{t+1|t}| \right) $.
We can remove $ln$ as $ln(x)$ is monotonically increasing for $x >0$.
Thus minimizing $x$ is equivalent to minimizing $ln(x)$. 
$\Sigmab_{t+1|t}$ is a constant at the given time step and the only term we can impact with our actions is $K_{t+1}$. 
Additionally $ 0 \leq |I - K_{t+1}\beljacob_{t+1}| \leq 1$, making this term easier to weight in an unconstrained optimization. 
As a result we choose Eq. \ref{eqn:true_entropy_loss} in place of entropy when implementing our entropy based loss. 

\begin{table}[t]
\setlength\tabcolsep{1em}
\centering
\caption{Additional comparison of estimation accuracy in simulated cases between BANet, the Adam optimizer, and the proposed method. Only one lift-up action is applied to all tissue because BANet only handles a single input.}
\begin{adjustbox}{width=0.5\textwidth}
    \begin{tabular}{l|ccc|ccc}
    \toprule
    & \multicolumn{3}{|c}{PCD \upgreenarrow}& \multicolumn{3}{|c}{PUG \upgreenarrow}\\ \midrule
 Cases &  BANet & Adam & Ours &  BANet & Adam & Ours\\ 
  \midrule
Arc & 10.7& 75.0& 100.0& 37.0& 5.6& 14.8\\
Arc-line& 9.4& 75.0& 100.0& 18.2& 5.5& 16.4\\
Line& 2.5& 75.0& 100.0& 16.7& 10.0& 16.7\\
Line-dot& 6.4& /& 100.0& 13.3& 0.0& 5.1\\
U-shape & 2.3& 80.0& 100.0& 4.8& 11.5& 22.1\\
    \bottomrule
\end{tabular}
\end{adjustbox}
\label{tbl:appendix_additional_estimation}
\end{table}

\begin{figure*}[t]
    \centering
    \includegraphics[width=1\linewidth]{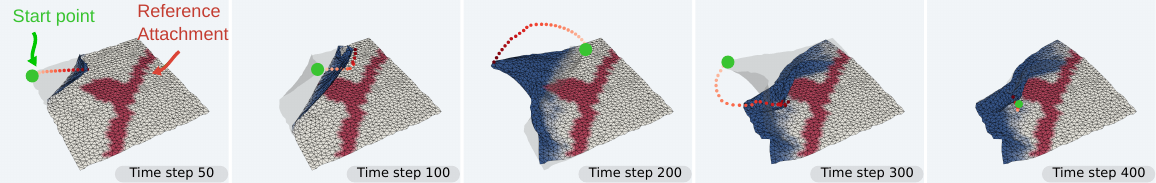}
    \caption{Active sensing performance of our framework with doubled amount of mesh particles (1200) on test cases \textit{Line-dot}. Our framework can still perform complex maneuvers such as switching directions and flipping for informaiton gain even with more particles, showing that it can generalize to complex meshes.}
    \label{fig:app_scalability}
\end{figure*}

\subsection{Additional Estimation Results}
\textbf{BANet \cite{9359348} Adaption and Comparison.} \label{appendix:banet_details}
In the main text, we only compared against previous method BANet on the real world data. 
Here, we provide additional information about how BANet was adapted for evaluation against our framework and also show additional comparisons against BANet on simulation data. 

BANet takes in a 3D volume representing the tissues and its deformation field at the end of each test cases as inputs. It then classifies if each voxel is boundary or not. We convert our triangular mesh to be volumetric by marking the closest voxel to each of our mesh particles to be part of the volumetric representation. \textcolor{rebuttal}{The mesh's particle displacement can be calculated with our proposed method (i.e. $\mathbf{x}^{\text{ref}}_{\text{end}} - \mathbf{x}^{\text{ref}}_0$).}
\textcolor{rebuttal}{\st{Assuming knowledge of mesh particles' displacement,}} A displacement field is computed by assigning the mesh particles' displacement to the closest voxel and setting the displacement everywhere else to be zero. \textcolor{rebuttal}{If the network output (after a sigmoid operation) on a voxel is larger than a threshold, we treat it as a boundary.}
\textcolor{rebuttal}{\st{We convert the network output (after a sigmoid operation) to binary boundary representations with a threshold of $0.1$.}}
We assign the threshold to be small ($0.1$) to include more relatively low-scored results from BANet because it assumes hard boundary conditions whereas our scenarios do not rely on that specific assumption. 
Lastly, boundary condition of mesh particles are determined by checking the BANet outputs corresponding to their closest voxel.

Additional comparisons between our method and BANet are shown in Table \ref{tbl:appendix_additional_estimation}. Looking at the PCD metric, BANet predicts inaccurate results. Fig.~\ref{fig:app_banet_compare} explanins it by showing that BANet tends to regard not-displaced particles as boundaries. 
In fact, if we increase the binary threshold to $0.5$, BANet predicts few boundary conditions. 
Note that this is correct for BANet because it assumes hard boundary conditions in its training data whereas in our test cases there is no hard boundary condition. 
But it also showed that BANet does not generalize to our attachment scenario that is common in surgery.

\subsection{Additional Active Sensing Results}
\textbf{Motion Primitive Details.} 
We show additional details from the experiment discussed in the paper when comparing our method with the motion primitives in \cite{boonvisut2014identification}. 
Due to ground contact, moving downward is forbidden. 
Therefore, we only consider 5 motion primitives of \textit{up}, \text{forward}, \textit{backward}, \textit{left}, and \textit{right}. There direction w.r.t to the tissue is shown in the left of Fig.~\ref{fig:app_active_sensing_pmps}. 
To avoid sliding, all primitives are preceeded by a grasp sequence that lifts up the tissue. Fig.~\ref{fig:app_active_sensing_pmps} shows additional comparison between the proposed method and all choices of predefined motion primitives (PMP).
All PMPs result in safety violation. 
Before violations, \textit{up} and \textit{forward} achieve more entropy reduction, but \textit{up} achieves that faster in general. 
In comparsion, our proposed method outperforms all PMPs, achieving more entropy reduction at the cost of smaller boundary energy.

\subsection{Scalability}
We evaluate our framework's ability to handle higher dimensional meshes. 
This become important when simulating surgical scenes of higher complexity. Specifically, we double the number of particles on our mesh (1200 particles). 
We show the performance of our framework with our most computation-demanding task, active sensing, in Fig.~\ref{fig:app_scalability}. Our framework is able to maintain the similar amount of information gain with more particles. 
The computation time of our active sensing controller increases from 2.9s to 4.5s. 
Despite increase in computation time, we believe our algorithm can be accelerated via techniques such as asynchronous computation and multi-resolution meshes to handle more complex scenarios.